\documentclass{article}

\usepackage{arxiv}

\usepackage[utf8]{inputenc} % allow utf-8 input
\usepackage[T1]{fontenc}    % use 8-bit T1 fonts
\usepackage{hyperref}       % hyperlinks
\usepackage{url}            % simple URL typesetting
\usepackage{booktabs}       % professional-quality tables
\usepackage{amsfonts}       % blackboard math symbols
\usepackage{nicefrac}       % compact symbols for 1/2, etc.
\usepackage{microtype}      % microtypography
\usepackage{graphicx}
\usepackage{doi}
\usepackage{multirow}%
\usepackage{amsmath,amssymb,amsfonts}%
\usepackage{amsthm}%
\usepackage{mathrsfs}%
\usepackage[title]{appendix}%
\usepackage{textcomp}%
\usepackage{manyfoot}%
\usepackage{algorithm}%
\usepackage{algorithmicx}%
\usepackage{algpseudocode}%
\usepackage{listings}%
\usepackage{booktabs}
\usepackage[table,xcdraw]{xcolor}
\usepackage{multirow}

\title{Robust Periorbital Distance Prediction Enables Generalizable Classification of Oculoplastic and Craniofacial Diseases}
\author{
    George R.~Nahass \\
    Ophthalmology and Biomedical Engineering \\
    University of Illinois Chicago \\
    Chicago, IL, USA \\
    \texttt{gnahas2@uic.edu} \\
    \And
    Sasha~Hubschman \\
    Ophthalmology \\
    University of Illinois Chicago \\
    Chicago, IL, USA \\
    \And
    Jeffrey C.~Peterson \\
    Ophthalmology \\
    University of Illinois Chicago \\
    Chicago, IL, USA \\
    \And
    Ghasem~Yazdanpanah \\
    Ophthalmology \\
    University of Illinois Chicago \\
    Chicago, IL, USA \\
    \And
    Nicholas~Tomaras \\
    Ophthalmology \\
    University of Illinois Chicago \\
    Chicago, IL, USA \\
    \And
    Madison~Cheung \\
    Plastic and Reconstructive Surgery \\
    University of Illinois Chicago \\
    Chicago, IL, USA \\
    \And
    Alexander~Palacios \\
    Plastic and Reconstructive Surgery \\
    University of Illinois Chicago \\
    Chicago, IL, USA \\
  \And
    Kevin~Heinze \\
    Ophthalmology \\
    Cornell University \\
    New York, NY, USA \\
    \And
    Chad~Purnell \\
    Plastic and Reconstructive Surgery \\
    University of Illinois Chicago \\
    Chicago, IL, USA \\
    \And
    Pete~Setabutr \\
    Ophthalmology \\
    University of Illinois Chicago \\
    Chicago, IL, USA \\
    \And
    Ann Q.~Tran\thanks{Co-corresponding authors: annqtran@uic.edu, dyi9@uic.edu} \\
    Ophthalmology \\
    University of Illinois Chicago \\
    Chicago, IL, USA \\
    \texttt{annqtran@uic.edu} \\
    \And
    Darvin~Yi\footnotemark[1] \\
    Ophthalmology and Biomedical Engineering \\
    University of Illinois Chicago \\
    Chicago, IL, USA \\
    \texttt{dyi9@}
}

\renewcommand{\headeright}{}
\renewcommand{\undertitle}{}
\renewcommand{\shorttitle}{Periorbital distance prediction and disease classification}

\hypersetup{
pdftitle={A template for the arxiv style},
pdfsubject={q-bio.NC, q-bio.QM},
pdfauthor={David S.~Hippocampus, Elias D.~Striatum},
pdfkeywords={First keyword, Second keyword, More},
}

\begin{document}
\maketitle

\begin{abstract}
Periorbital distances are essential for diagnosing and monitoring oculoplastic and craniofacial conditions, but manual measurements are subjective and susceptible to intergrader variability. While automated approaches have emerged, they remain limited by reliance on standardized images, small datasets, and a narrow scope of measurable features. This study has two primary objectives. First, we sought to improve the accuracy of automated periorbital measurements in both healthy and diseased eyes. We developed a segmentation pipeline trained on a domain-specific dataset of healthy eyes and benchmarked it against the Segment Anything Model (SAM) and PeriorbitAI. Segmentation accuracy was evaluated across multiple disease classes and imaging conditions. Second, we assessed whether periorbital distances derived from these models could support disease classification under both in-distribution (ID) and out-of-distribution (OOD) settings, using shallow classifiers, convolutional neural networks (CNNs), and fusion models. Our model achieved state-of-the-art segmentation accuracy, with error rates within intergrader variability and superior performance compared to SAM and PeriorbitAI. For classification, periorbital distance-based models matched CNNs in ID settings (77–78\% accuracy) and significantly outperformed them under OOD conditions (63–68\% vs. 14\%). Fusion models reached the highest ID accuracy (80\%) but were sensitive to CNN degradation under domain shift. These results establish a new benchmark for periorbital distance prediction and demonstrate that segmentation-derived anatomical features offer a robust, interpretable foundation for AI-driven disease classification in real-world clinical settings.
\end{abstract}

\section{Introduction}

Periorbital distances are critical to diagnose and monitor a range of oculoplastic and craniofacial conditions. For example, diseases like ptosis, thyroid eye disease (TED), Apert syndrome, Crouzon syndrome, and other craniofacial disorders are characterized by distinct periorbital features \cite{lootus_development_2023,kabak_hemifacial_2019,chetty_improvement_2020,strianese_methotrexate_2014}, which are measured manually and used to monitor disease progression as well as evaluate the effects of surgical or medical interventions \cite{boboridis_repeatability_2001,bhalodia_quantifying_2020,lou_novel_2019}. However, manual measurements are time-consuming, subjective, and prone to substantial intergrader variability \cite{boboridis_repeatability_2001,van_brummen_periorbitai_2021}. Leveraging artificial intelligence (AI) offers a promising avenue for standardizing these measurement, improving accuracy, and increasing reproducibility across different providers and clinical settings \cite{lootus_development_2023,lou_novel_2019,van_brummen_periorbitai_2021,cao_novel_2021,fu_artificial_2023,rana_artificial_2024,chen_smartphone-based_2021}.

Despite recent advances, current AI pipelines face significant limitations. Van Brummen et al. and Rana et al. trained models to collect periorbital measurements , but these were developed on small datasets of standardized images captured in controlled settings \cite{van_brummen_periorbitai_2021,rana_artificial_2024}. Chen et al.’s smartphone-based approach, limited to ptotic eyelids and MRD1/MRD2 measurements, also required strict image acquisition protocols\cite{chen_smartphone-based_2021}. These models not only capture a narrow range of periorbital metrics but also lack validation in diverse imaging conditions that better reflect real-world clinical practice. In reality, patient photographs are often taken using various devices (e.g., cameras, smartphones) under inconsistent lighting and without standardized protocols. There remains a critical need for automated periorbital segmentation that performs reliably in these clinically relevant settings.

Moreover, while accurate periorbital measurements are clinically valuable, their potential as features for downstream tasks in AI models—such as disease classification—remains unexplored. Deep learning models have the potential to aid or even complement the diagnosis of oculoplastics and craniofacial disorders,  \cite{conrady2018apert,Zeppieri2025crouzon,marszalek2023children,barbosa2004treacher}. To effectively train deep learning models for the diagnosis of these diseases, large and diverse datasets of clinical images must be collected from multiple institutions. However, this introduces challenges related to generalizability, as domain shift and dataset bias can significantly affect model performance. \cite{quinonero2022dataset,stacke2019closer,pooch2019can,moreno2012unifying,rashidisabet2023validating}. As noted previously, clinical images are often captured under variable conditions—using different devices and lighting setups—without standardized protocols, making it difficult for models to learn robust features. To address this limitation, we hypothesize that using segmentation-derived periorbital measurements as input features, rather than raw clinical images, may enhance generalizability to “out-of-distribution” (OOD) scenarios, such as non-standardized imaging environments.

This study has two primary objectives. First, we sought to improve the accuracy of automated periorbital measurements in both healthy and diseased eyes. To achieve this, we developed a segmentation model (DeepLabV3) and evaluated its performance alongside other segmentation approaches, including the foundational gold-standard Segment Anything Model (SAM), benchmarking their accuracy on diverse clinical image datasets. Second, we investigated whether periorbital distances derived from these segmentation models could outperform current state-of-the-art methods in diagnosing oculoplastic and craniofacial disorders, across both in-distribution (ID) and out-of-distribution (OOD) datasets. Our findings demonstrate that these extracted periorbital measurements not only rival or exceed the performance of existing models, but also provide robust and generalizable features for disease classification. To our knowledge, this is the first study to establish benchmarks for periorbital distance prediction and to systematically evaluate their diagnostic utility under distribution shift.

\section{Methods}

\begin{table}[]
\centering
\resizebox{\columnwidth}{!}{%
\begin{tabular}{@{}|c|c|c|c|@{}}
\toprule
\textbf{Dataset}                    & \textbf{Num. Images} & \textbf{Source}     & \textbf{Utility} \\ \midrule
\textbf{Syndrome}                   & 69                & UIC-CFC             & Testing          \\ \midrule
\textbf{Facial Asymmetry}           & 12                & UIC-CFC             & Testing          \\ \midrule
\textbf{Craniosynostosis}           & 11                & UIC-CFC             & Testing          \\ \midrule
\textbf{Cleft}                      & 13                & UIC-CFC             & Testing          \\ \midrule
\textbf{Unknown/Other Craniofacial} & 50                & UIC-CFC             & Testing          \\ \midrule
\textbf{Thyroid Eye Disease}        & 113               & UIC-Oph             & Testing          \\ \midrule
\textbf{Healthy}                    & 827               & Open Source (CFD)   & Testing          \\ \midrule
\textbf{Celeb}                      & 2015            & Open Source (Celeb) & Training         \\ \bottomrule
\end{tabular}%
}
\caption{Details of all imaging datasets used in this study for training and testing segmentation model. ‘Num. Images’ denotes the number of images used in evaluating the various pipelines, ‘Source’ indicates where the data was acquired from, and ‘Utility’ indicates the role of the dataset in our models. DeepLabV3 models were trained on the CelebAMask dataset, but these images were not included in any periorbital distance evalaution experiments \cite{lee_maskgan_2020}.}
\label{tab:tab-1}
\end{table}

\subsection{Datasets}

In the following sections, we describe the datasets used for training and evaluating segmentation and classification models.

\subsubsection{Segmentation and Periorbital Distance Datasets}

To train the DeepLabV3 segmentation model, we used 2,015 images from an open-source periorbital segmentation dataset annotated for iris, lid, caruncle, sclera, and brow \cite{nahass2025open}. For evaluation of periorbital distance predictions, we used 827 images from the open-source Chicago Facial Dataset (CFD) as healthy controls, 113 images of patients with Thyroid Eye Disease (TED) from the University of Illinois at Chicago (UIC) Ophthalmology Clinic, and 155 images from the UIC Craniofacial Center comprising various craniofacial syndromes (Table \ref{tab:tab-1}) (\cite{ma_chicago_2015,ma_chicago_2021,lakshmi_india_2021}). A full breakdown of these evaluation datasets is available in the supplement (Supplemental Table \ref{tab:sup_big_bucket}).

Craniofacial images were acquired by a trained photographer, while TED images were captured using smartphones during routine clinic visits. All images were aligned such that the line from the nasion to the hairline midpoint was vertical.

Ground truth segmentation masks for periorbital distance evaluation were manually generated using the Computer Vision Annotation Tool (CVAT) by trained annotators. For scleral boundaries, annotations extended from the lateral to medial canthus and included the caruncle. Ground truth periorbital distances were computed using the same anatomical measurement pipeline applied to AI-generated masks. Examples of annotated masks and extracted measurements are shown in Supplemental Figure \ref{fig:fig_1}. Intergrader variability was assessed using five annotators on a shared subset of 100 images, with standard deviation used to quantify measurement variance (Supplemental Figure \ref{fig:grader-schematic}).

\subsubsection{Classification Datasets}

For classification experiments, we assembled a 13-class dataset of 2,742 images derived from the UIC Craniofacial Center and UIC Ophthalmology Clinic \cite{nahass2024facefinder}. This served as our in-distribution (ID) dataset. An out-of-distribution (OOD) dataset was created by scraping publicly available web images for a subset of the conditions.

The ID dataset included the following classes: healthy adult, healthy pediatric, Crouzon-Apert-Pfeiffer (CAP), Goldenhar syndrome, facial asymmetry, fibrous dysplasia, craniosynostosis, miscellaneous or unclassified syndromes, Nager syndrome, Parry Romberg syndrome, ptosis, TED, and Treacher Collins syndrome. Age was the only clinical variable collected for all subjects, as we observed pediatric status being a large predictive feature for the classification models. Table \ref{tab:datasets_class} provides the full class breakdown and data sources for both ID and OOD samples.

\begin{table}[]
\centering
\resizebox{\columnwidth}{!}{%
\begin{tabular}{@{}|c|c|c|c|c|c|c|c|c|c|c|c|c|c|c|@{}}
\toprule
\cellcolor[HTML]{FFFFFF} &
  Disease &
  CAP &
  Craniosynostosis &
  Facial Asymmetry &
  Fibrous Dysplasia &
  Goldenhar &
  Healthy Adult &
  Healthy Ped. &
  Misc. Syndrome &
  Nager Syndrome &
  Parry Romberg &
  Ptosis &
  TED &
  Treacher Collins \\ \midrule
                      & Count:  & 129 & 379 & 144 & 26  & 317 & 826 & 389 & 111 & 35  & 46  & 80  & 198 & 62  \\ \cmidrule(l){2-15} 
\multirow{-2}{*}{ID} &
  Source: &
  UIC-CFC &
  UIC-CFC &
  UIC-CFC &
  UIC-CFC &
  UIC-CFC &
  CFD &
  UIC-CFC &
  UIC-CFC &
  UIC-CFC &
  UIC-CFC &
  UIC-OPH &
  UIC-OPH &
  UIC-CFC \\ \midrule
                      & Count:  & 18  & 0   & 0   & 0   & 6   & 20  & 0   & 0   & 0   & 0   & 13  & 20  & 11  \\ \cmidrule(l){2-15} 
\multirow{-2}{*}{OOD} & Source: & WEB & WEB & WEB & WEB & WEB & WEB & WEB & WEB & WEB & WEB & WEB & WEB & WEB \\ \bottomrule
\end{tabular}%
}
\caption{Details on datasets for classification used in this study. Abbreviations are as follows: CAP- Crouzon, Apert, and Pfeiffer syndrome which have similar ophthalmic presentations, TC-Treacher Collins, UIC-CFC- UIC Craniofacial Center, UIC-O- UIC Ophthalmology, Web- images scraped from the web, CFD- Chicago Facial Dataset.}
\label{tab:datasets_class}
\end{table}

\subsection{Periorbital Distance Prediction Pipelines}

The following two sections describe our approaches to predicting periorbital distances through a segmentation network (Figure \ref{fig:overview}A).

\subsubsection{DeepLabV3 }

\begin{figure}[t]
    \centering
    \includegraphics[width=1\linewidth]{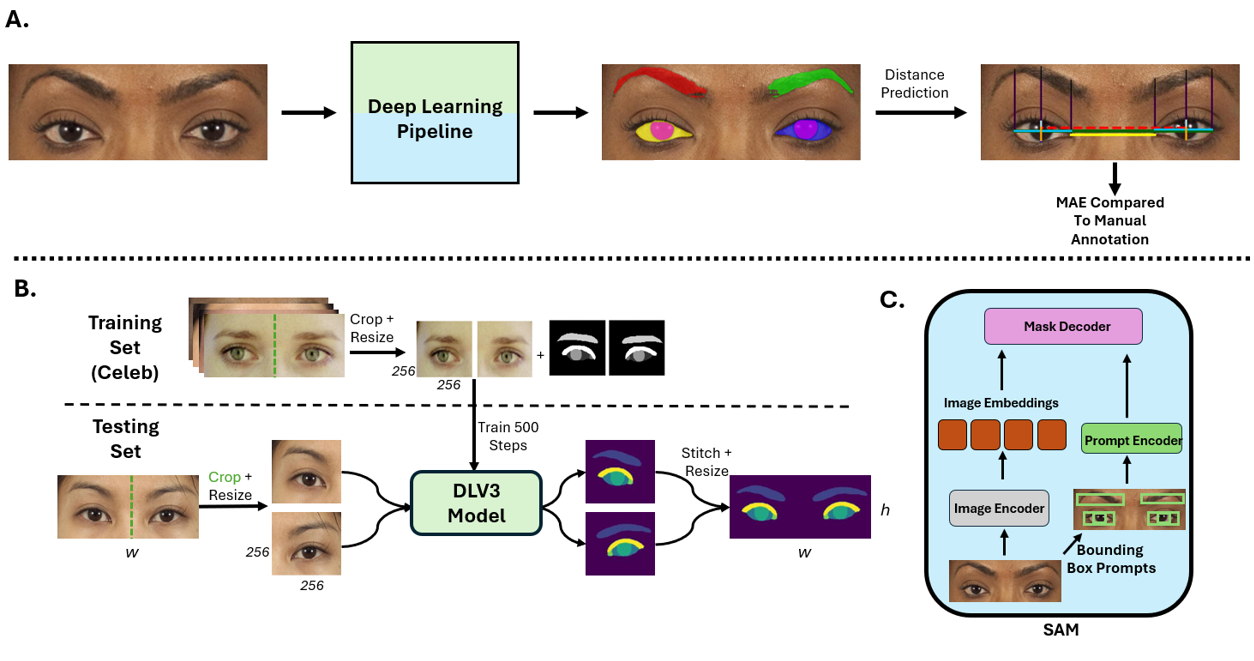}
\caption{ Graphical schematic of segmentation models and distance prediction pipelines. A) Cropped images of the eyes are segmented using one of two deep learning
models (DLV3 or SAM). Following segmentation, the Dice score was calculated, and the
segmentation masks are used to predict periorbital distances which were compared to distances
obtained from human annotations using the mean absolute error (MAE). B) Graphical schematic of the training procedure for the DLV3 model, Input images are cropped at the midline and both eyes are resized to be $256x256$. The network is then trained for 500 steps. Details can be found in the methods. C) Graphical schematic of the Segment Anything Model (SAM). The cropped image
was used as input, and bounding box prompts were derived from MediaPipe facemesh
coordinates \cite{kartynnik_real-time_nodate}.}
    \label{fig:overview}
\end{figure}

 For training a DeepLabV3 segmentation model, each image was split at the midline to isolate the left and right eye, and both halves were resized to 256×256 pixels. A DeepLabV3 model was trained using a custom-labeled dataset specifically designed for periorbital applications, which included annotations for the eyelids, brows, iris, and caruncle (\cite{nahass2025open}). The network was trained for 500 steps using cross-entropy loss and the Adam optimizer with a learning rate of 0.01 and $\beta_1 = 0.5$, $\beta_2 = 0.99$. The resulting segmentation masks were used to predict periorbital distances through standard computer vision techniques described below. A schematic of the DeepLabV3 training and inference pipeline is shown in Figure \ref{fig:overview}B.

% Segmentation quality was evaluated using the Dice score (Equation \ref{dice}). 

% \begin{equation}
%     \label{dice}
%     Dice = \frac{2(\mathbf{X}\cap{\mathbf{Y}})}{|\mathbf{X}| + |\mathbf{Y}|}
% \end{equation}

\subsubsection{Segment Anything Model}
A 468-point face mesh was fit to each input image using MediaPipe \cite{kartynnik_real-time_nodate}. Using the facial landmark coordinates, bounding boxes were generated for the left and right sclera, brows, and irises. Each image was also cropped around the ocular region to improve segmentation focus. These bounding boxes and cropped regions were submitted to the Segment Anything Model (SAM) using the open-source \texttt{vit\_h} weights (Figure \ref{fig:overview}C) \cite{kirillov_segment_2023}. The resulting segmentation masks were used to compute periorbital distances using standardized computer vision techniques described below. A schematic of the complete pipeline is shown in Figure \ref{fig:overview}.

\subsection{Calculation of Anatomical Relationships}
 The iris diameter was standardized to 11.71 millimeters (mm), which was used to derive pixel-to-mm conversions as described by Van Brummen et al. \cite{van_brummen_periorbitai_2021}. Inferior and superior scleral show were measured as the vertical distance from the inferior and superior boundaries of the iris to the respective eyelid margins. Margin Reflex Distances 1 and 2 (MRD 1 and MRD 2) were computed as the distance from the center of the iris to the upper and lower eyelid, respectively.

Inner canthal distance (ICD), outer canthal distance (OCD), and interpupillary distance (IPD) were defined as the horizontal distances between the medial canthi, lateral canthi, and iris centers, respectively. Brow heights were calculated as the vertical distance from the medial canthus, lateral canthus, and iris center to the nearest brow point sharing the same $x$ coordinate.

Canthal tilt was defined as the angle between two lines: one running from the medial to the lateral canthus, and another extending vertically from the medial canthus (defined by the soft tissue nasion and midpoint of the hairline). Vertical dystopia was computed as the Euclidean distance between the left and right medial canthi projected onto this vertical facial axis. Medial and lateral canthal heights were defined as the vertical distances from each canthus to the line connecting the iris centers.

The vertical palpebral fissure was computed as the sum of MRD 1 and MRD 2. The horizontal palpebral fissure was defined as the horizontal distance between the medial and lateral canthus. For bilateral measurements, the difference between left and right was calculated to assess asymmetry. Scleral area was calculated as the ratio of the iris area to the scleral segmentation mask. A labeled diagram of all periorbital distances is shown in the supplemental materials (Figure \ref{fig:fig_1}).

\subsection{Comparison to PeriorbitAI}
The open-source code and pretrained weights for PeriorbitAI were obtained for benchmarking purposes. The source code was modified so that all measurements defaulted to zero in the event of failed predictions, allowing the pipeline to continue execution without interruption. As described in the original publication \cite{van_brummen_periorbitai_2021}, all testing images were cropped to include the nose and forehead and resized using bilinear interpolation prior to inference.

PeriorbitAI was then applied to all testing datasets (Table \ref{tab:tab-1}). Anatomic measurements were computed using the same procedures described in Van Brummen et al., ensuring consistency across comparisons. For each measurement, we recorded the proportion of images that failed to yield a valid output. To ensure fairness, mean absolute error (MAE) for both our models and PeriorbitAI was computed only on images where PeriorbitAI produced a valid prediction (Equation \ref{mae}). For example, if PeriorbitAI failed to compute right MRD 1 for a particular image, that image was excluded from the MAE calculation for both models. This approach ensured that comparison metrics were based on identical subsets of evaluable cases.

\begin{figure}[t]
    \centering
    \includegraphics[width=\textwidth]{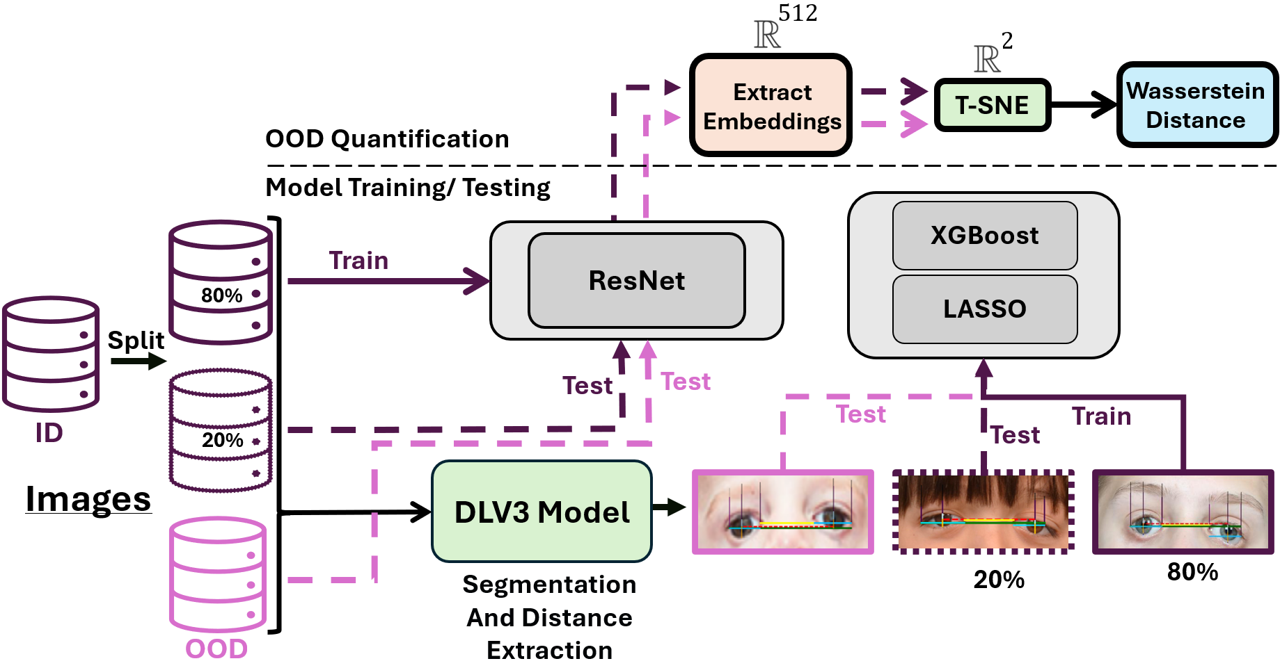}
    \caption{ Graphical overview of the classification pipeline used in this study. Models were trained on ID data and tested on both ID and OOD datasets. XGBoost and Lasso models were trained using periorbital distances extracted via a segmentation (DeepLabV3) intermediate step  (bottom). A ResNet-18 was trained for classification on cropped images. OOD-ness was quantified (top) by inspecting the embeddings produced by the trained ResNet.}
    \label{fig:ml_schem}
\end{figure}

\begin{equation}
    \label{mae}
   MAE = \frac{1}{n}\sum_{i=1}^n|\hat{y}_i-y_i|
\end{equation}

\subsection{Classification Pipelines}
We evaluated multiple classification strategies using either raw images, engineered periorbital distances, or a combination of both. All models were trained exclusively on in-distribution (ID) data and evaluated on both ID and out-of-distribution (OOD) datasets. A schematic overview of the classification pipeline is shown in Figure \ref{fig:ml_schem}.

\subsubsection{CNN Baselines}
A convolutional neural network was trained for multiclass disease classification using a ResNet-18 architecture pretrained on ImageNet, a large standardized image pretraining dataset \cite{he2015deep}. The network was fine-tuned on cropped images of the eyes. 

Networks were trained using cross-entropy loss and optimized with the Adam optimizer (learning rate 0.001, batch size 16) for 20 epochs. The model checkpoint with the best validation accuracy was selected for downstream testing. Performance was evaluated using top-1 and top-3 accuracy, F1 score, and area under the receiver operator characteristic curve (AUROC) on both ID and OOD data. 

To evaluate model behavior, t-SNE was applied to the penultimate layer embeddings of the CNN to visualize latent space organization and assess class separability and domain shift between ID and OOD samples \cite{rashidisabet2023validating}. To quantify distributional differences between ID and OOD datasets, Wasserstein distances were computed for both CNN-derived embeddings and tabular periorbital distance features. Grad-CAM was used to generate saliency maps for CNN predictions. 

\subsubsection{Shallow Models with Periorbital Distances}

Two machine learning models—Lasso and XGBoost—were trained using periorbital distances extracted from segmentation masks generated by a DeepLabV3 model. All features were extracted from the ID dataset, and labels were stratified during an 80/20 train-test split to preserve class balance. Hyperparameters were chosen using grid search and 5-fold cross-validation on the training set. 

Lasso was implemented as a logistic regression model with L1 regularization, using the SAGA solver with one-vs-rest (OVR) multiclass handling ($C=1.0$, max iterations = 10,000). XGBoost was trained as a multiclass classifier using the softmax objective. The best-performing configuration, identified by cross-validated accuracy, used 300 estimators, a learning rate of 0.01, and a maximum tree depth of 7. Feature importance was also extracted from the trained XGBoost model to identify the most predictive periorbital distances.

\subsubsection{Fusion Models}
To evaluate whether CNN and tabular features capture complementary information, we implemented a late-fusion model combining ResNet-extracted image embeddings and periorbital distances. Embeddings from the penultimate layer of the ResNet were concatenated with normalized periorbital features and passed to a multi-layer perceptron (MLP) for classification. The MLP was trained on the ID dataset and evaluated on both ID and OOD test sets.

\subsection{Hardware and Statistical Analysis}
All deep learning experiments were performed on three NVIDIA 1080Ti GPUs. Classical machine learning models and statistical analyses were conducted on an AMD Ryzen 7 7800x3D CPU. All code was implemented in Python 3.8.

Mean absolute error (MAE) for periorbital measurements was calculated according to Equation \ref{mae}. Bland–Altman plots were used to assess agreement between model-predicted and human-annotated distances. For each measurement, the mean difference and 95\% limits of agreement (mean ± 1.96 standard deviations) were computed. The percentage of samples falling outside the limits of agreement was also reported.

For classification tasks, performance was evaluated using accuracy, precision, recall, F1 score, and AUROC (Equation \ref{ml_metric}). Confidence intervals for AUROC were estimated using 1000-sample bootstrap resampling with replacement. Only resampled sets containing all classes were retained. All statistical analyses were conducted in Python.

\begin{equation}
    \label{ml_metric}
   \text{Acc} = \frac{TP + TN}{TP + TN + FP + FN}, \quad 
   \text{Pr} = \frac{TP}{TP + FP}, \quad 
   \text{Re} = \frac{TP}{TP + FN}, \quad
   \text{F1} = \frac{2 \cdot \text{Pr} \cdot \text{Re}}{\text{Pr} + \text{Re}}
\end{equation}

\section{Results}\label{results}

\subsection{Distance Prediction Accuracy}

A DeepLabV3 segmentation model was trained to convergence and used alongside the Segment Anything Model (SAM) to predict periorbital distances. Predictions were evaluated at two levels: (1) pixel-wise agreement using Bland–Altman plots, and (2) millimeter-level accuracy after scaling by iris diameter.

\subsubsection{Pixel-Level Evaluation}

\begin{figure}[H]
    \centering
    \includegraphics[width=1\linewidth]{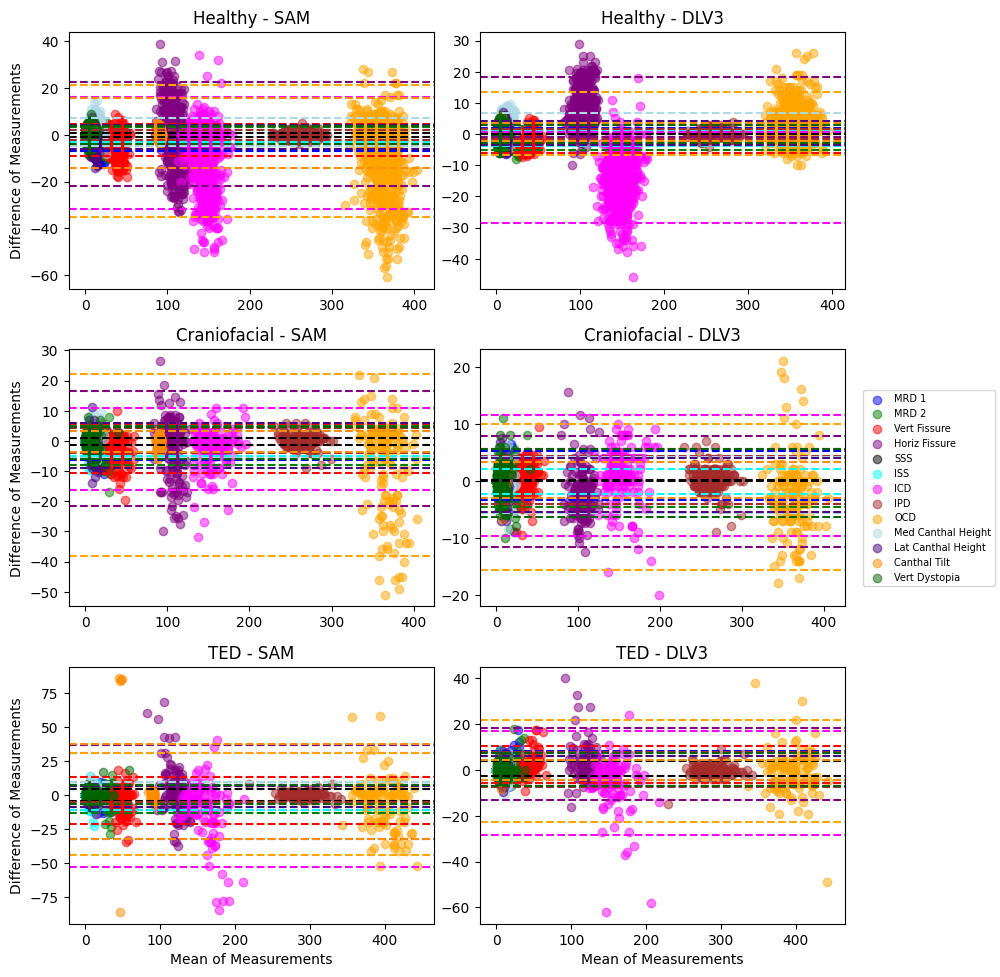}
    \caption{Bland Altman plots for the predicted eye distances from the human
annotated images and from all of our models for all of our datasets. These plots were generated using the pixel level distances. Bilateral distances were averaged, and dashed lines represent the 95\% limits of agreement (mean difference ± 1.96 standard deviations). Abbreviations can be interpreted as follows: MRD-Margin to Reflex Distance, ICD-inner canthal distance, IPD- interpupillary distance, OCD- outer canthal distance, ISS- inferior scleral show, SSS- superior scleral show, Vert Fissure- vertical palpebral fissure, Horiz Fissure- horizontal palpebral fissure, Med Canthal- medial canthal height, Lat Canthal- lateral canthal height.}
    \label{fig:eye_ba}
\end{figure}

\begin{figure}[H]
    \centering
    \includegraphics[width=1\linewidth]{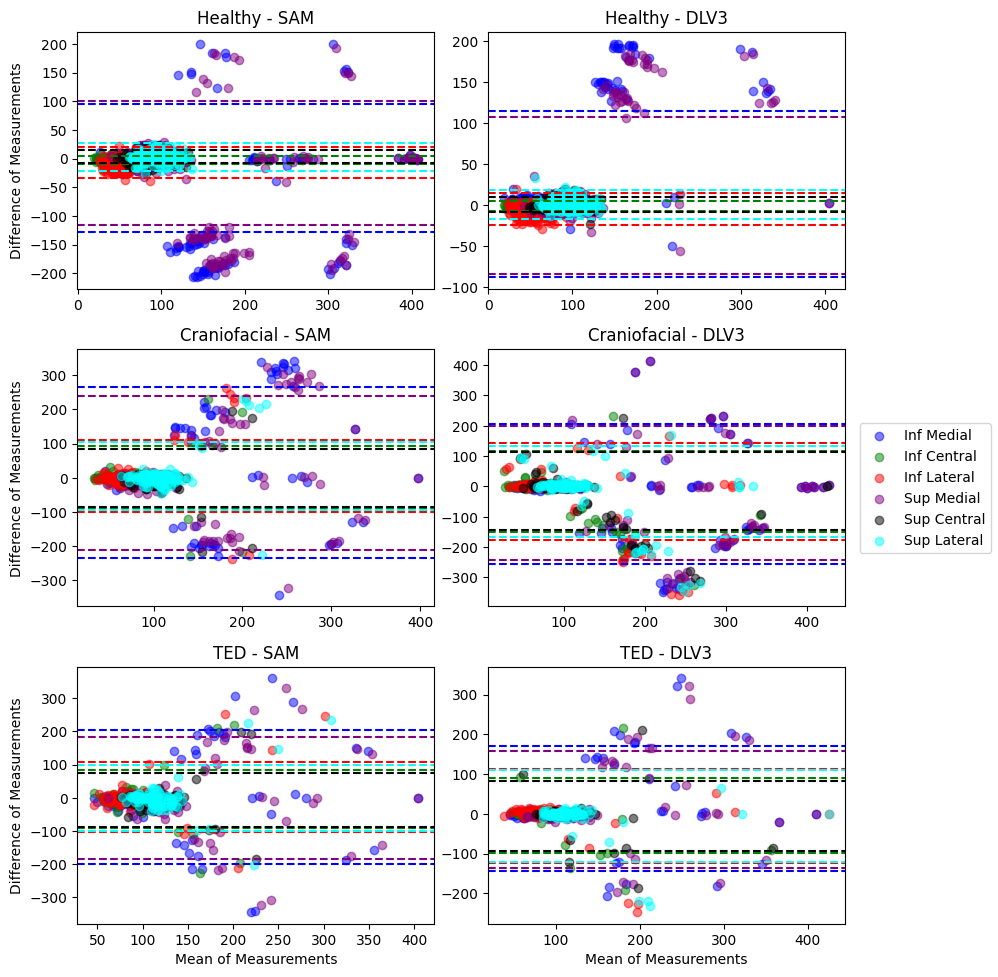}
    \caption{Bland Altman plots for the predicted brow distances from the human
annotated images and from all of our models for all of our datasets. These plots were generated using the pixel level distances. Bilateral distances were averaged, and dashed lines represent the 95\% limits of agreement (mean difference ± 1.96 standard deviations). Abbreviations can be interpreted as follows: Sup. Medial- superior medial brow height, Sup. Central- superior central brow height, Sup. Lateral- superior lateral brow height, Inf. Medial- inferior medial brow height, Inf. Central- inferior central brow height, Inf. Lateral- inferior lateral brow height.}
    \label{fig:brow_ba}
\end{figure}

Bland–Altman plots were created on the averaged bilateral measurements to compare AI-predicted and human-annotated distances (Supplemental Figures \ref{fig:eye_ba}, \ref{fig:brow_ba}). DeepLabV3 yielded the most accurate results, with the lowest percentage of eye measurements outside the limits of agreement across all datasets—4.63\% compared to 5.37\% for SAM. Full results for eye and brow measurements are provided in Supplemental Tables \ref{tab:supp_eye_ba} and \ref{tab:supp_brow_ba}.

\subsubsection{Millimeter-Level Evaluation}

Across all disease classes and measurements, DeepLabV3 yielded substantially lower MAEs than SAM, with an average improvement of $0.8$ mm (Table \ref{tab:mae}). The largest gains were observed in TED and syndromic cases, where SAM often produced less anatomically accurate segmentation masks, leading to greater variability in distance estimation. The only exception was outer canthal distance (OCD) in healthy controls, where SAM slightly outperformed DeepLabV3 ($2.89$mm vs. $4.41$mm). However, DeepLabV3 achieved lower OCD errors across all other disease groups (Table \ref{tab:mae}).

\begin{figure}[H]
    \centering
    \includegraphics[width=1\linewidth]{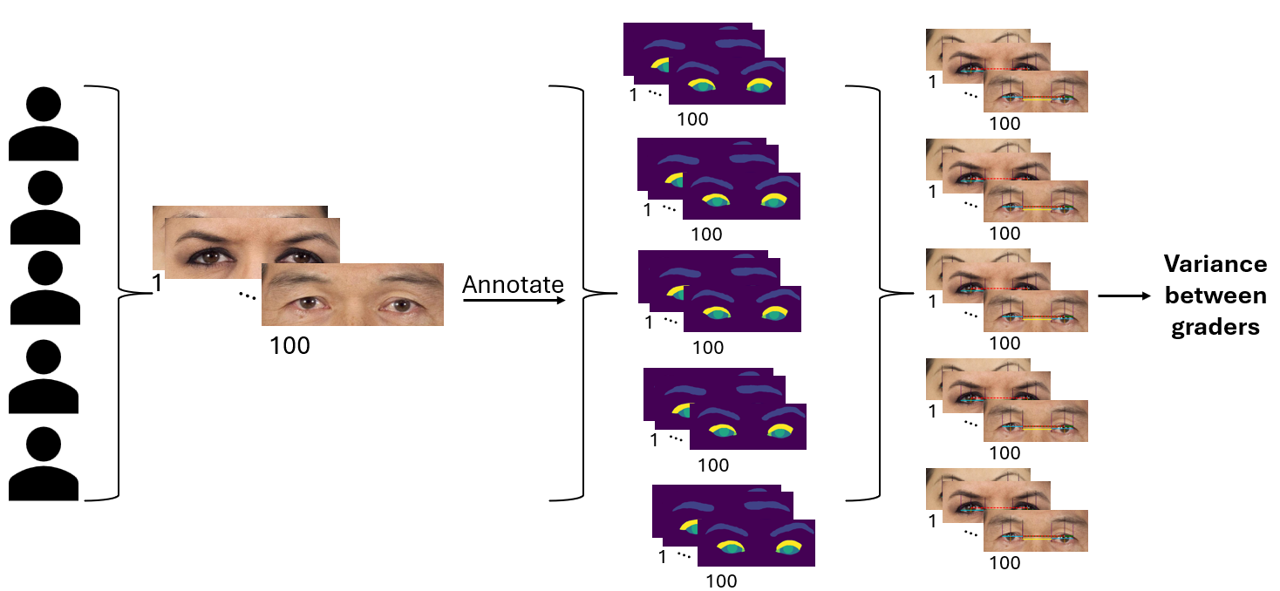}
    \caption{Graphical schematic of process used to establish intergrader variance on key periorbital measurements. Five annotators were asked to annotate the same 100 images of healthy eyes. Periorbital distances were computed for all images from all graders, and the standard deviation of each meaurement was calculated.}
    \label{fig:grader-schematic}
\end{figure}

Previously reported intergrader variability ranges up to 0.5 mm for MRD 1/MRD 2 and 4 mm for ICD/OCD \cite{boboridis_repeatability_2001, van_brummen_periorbitai_2021}, which our analysis reproduced (Supplemental Figure \ref{fig:grader-schematic}, Table \ref{tab:eye_std}). In most cases, DeepLabV3 MAEs were below these intergrader thresholds. MRD 1, MRD 2, and ICD were consistently below the threshold, while 86\% of ISS, SSS, and horizontal fissure measurements fell within threshold limits. Outer canthal distance (OCD) was the most difficult to predict, with none of the MAEs falling below the intergrader threshold. Nonetheless, DeepLabV3 consistently outperformed SAM (Supplemental Figure \ref{fig:threshold}). Detailed MAE values are provided in Table \ref{tab:mae}, with error distributions shown in Supplemental Figure \ref{fig:eye_histos}. Examples of predicted periorbital distances from both models on representative images of various disease classes can be seen in Figure \ref{fig:qual}

\begin{figure}[H]
    \centering
    \includegraphics[width=1.\linewidth]{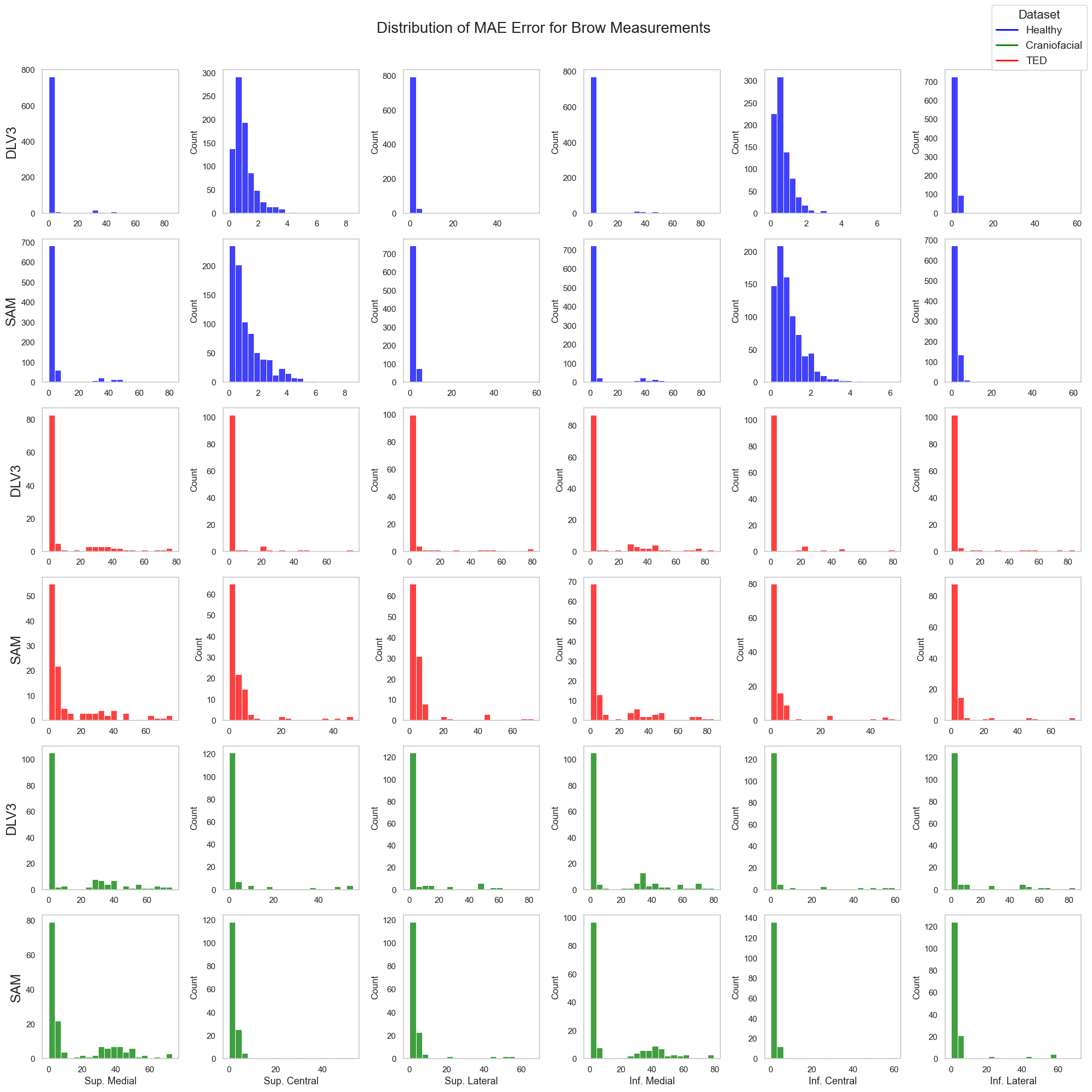}
    \caption{Histograms of MAE for all brow measurements using all models for all ndatasets. Bilateral distances were averaged. Scale is in mm following conversion using the iris as a diameter.}
    \label{fig:brow_histos}
\end{figure}

\begin{figure}[H]
    \centering
    \includegraphics[width=1\linewidth]{ 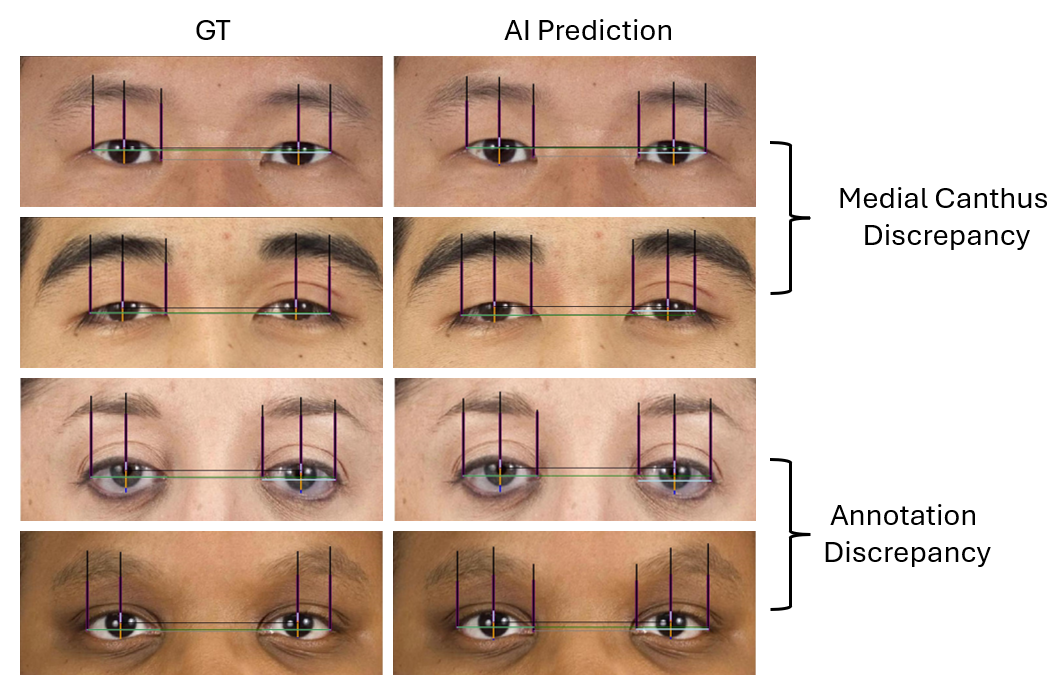}
    \caption{Representative examples of brow height discrepancies resulting from either human annotator judgment of medial brow margin or insufficient medial scleral segmentation.}
    \label{fig:brow_discrep}
\end{figure}

Brow height MAEs are listed in Supplemental Table \ref{tab:supp_brow_mae}. The distribution was long-tailed (Figure \ref{fig:brow_histos}) due to labeling inconsistencies and occasional segmentation errors (Figure \ref{fig:brow_discrep}). After excluding outliers >1 standard deviation, recalculated MAEs were consistent with intergrader variability measurements (Supplemental Table \ref{tab:brow_std} and Supplemental Figure \ref{tab:supp_brow_mae_o}). Representative examples of eye and brow predictions are shown in Figure \ref{fig:qual}.

\subsubsection{Comparison to PeriorbitAI}

We compared our pipelines to PeriorbitAI, the current state-of-the-art (SOTA) method for periorbital distance prediction \cite{van_brummen_periorbitai_2021}. PeriorbitAI failed to process the entire dataset. The average proportion of images successfully analyzed was 85\% for Healthy, 59\% for TED, and 84\% across craniofacial disease datasets whereas our approach analyzed all of the images sucessfully.

Performance comparisons were made using only the subset of images successfully processed by PeriorbitAI. On these, our pipelines outperformed PeriorbitAI on all but two measurements: outer canthal distance (OCD) in TED, and MRD 2 in Facial Asymmetry (Table \ref{tab:pai}). For brow distances, our models outperformed PeriorbitAI on all but one (superior medial brow height in the Unknown/Other Craniofacial group; Supplemental Table \ref{tab:supp_brow_pai}).

\begin{figure}[t]
    \centering
    \includegraphics[width=.75\linewidth]{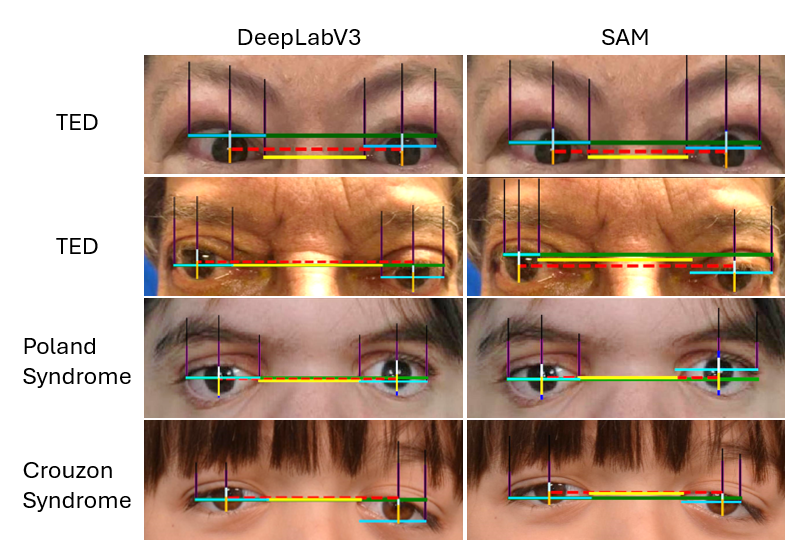}
    \caption{Qualitative evaluation of all three models periorbital distance prediction. Predicted distances from DeepLabV3 and SAM (left to right columns) for various disease states. Brightness has been increased on some images for presentation purposes only. Color can be interpreted as follows: Red dashes: IPD, teal: HPF, light blue: MRD 1, orange: MRD 2, green: OCD, yellow: ICD, purple: Inferior brow height, black: superior brow height.}
    \label{fig:qual}
\end{figure}

\begin{table}[]
\centering
\resizebox{\columnwidth}{!}{%
\begin{tabular}{@{}|c|c|c|c|c|c|c|c|c|@{}}
\toprule
\cellcolor[HTML]{FFFFFF}Disease   Class &
  Segmentation   Model &
  MRD 1 &
  MRD 2 &
  ICD &
  OCD &
  ISS &
  SSS &
  Horiz. Fissure \\ \midrule
 &
  SAM &
  0.61 ± 0.56 &
  0.44 ± 0.40 &
  2.51 ± 2.68 &
  \textbf{{\color{red}2.89 ± 3.17}} &
  0.19 ± 0.32 &
  0.02 ± 0.12 &
  2.34 ± 1.78 \\ \cmidrule(l){2-9} 
\multirow{-2}{*}{Healthy} &
  DLV3 &
  \textbf{0.32 ± 0.23} &
  \textbf{0.40 ± 0.28} &
  \textbf{1.49 ± 0.94} &
  4.41 ± 1.63 &
  \textbf{0.09 ± 0.15} &
  \textbf{0.00 ± 0.00} &
  \textbf{1.50 ± 0.83} \\ \midrule
 &
  SAM &
  0.91 ± 0.75 &
  1.12 ± 1.10 &
  4.33 ± 4.05 &
  5.20 ± 3.65 &
  0.74 ± 0.98 &
  0.22 ± 0.45 &
  3.42 ± 2.82 \\ \cmidrule(l){2-9} 
\multirow{-2}{*}{TED} &
  DLV3 &
  \textbf{0.39 ± 0.28} &
  \textbf{0.37 ± 0.24} &
  \textbf{1.11 ± 0.88} &
  \textbf{2.39 ± 1.81} &
  \textbf{0.15 ± 0.22} &
  \textbf{0.05 ± 0.12} &
  \textbf{1.15 ± 0.84} \\ \midrule
 &
  SAM &
  0.54 ± 0.42 &
  0.81 ± 0.77 &
  1.79 ± 1.5 &
  2.92 ± 2.01 &
  0.32 ± 0.7 &
  0.01 ± 0.04 &
  1.41 ± 1.14 \\ \cmidrule(l){2-9} 
\multirow{-2}{*}{Cleft Palate} &
  DLV3 &
  \textbf{0.21 ± 0.11} &
  \textbf{0.22 ± 0.13} &
  \textbf{0.97 ± 0.78} &
  \textbf{1.29 ± 1.11} &
  \textbf{0.00 ± 0.00} &
  \textbf{0.00 ± 0.00} &
  \textbf{0.47 ± 0.31} \\ \midrule
 &
  SAM &
  0.48 ± 0.25 &
  0.46 ± 0.32 &
  1.71 ± 1.35 &
  2.86 ± 2.18 &
  0.07 ± 0.15 &
  0.0 ± 0.0 &
  1.24 ± 0.97 \\ \cmidrule(l){2-9} 
\multirow{-2}{*}{Craniosynostosis} &
  DLV3 &
  \textbf{0.22 ± 0.09} &
  \textbf{0.27 ± 0.09} &
  \textbf{1.01 ± 0.46} &
  \textbf{1.80 ± 1.79} &
  \textbf{0.01 ± 0.03} &
  \textbf{0.00 ± 0.00} &
  \textbf{0.65 ± 0.45} \\ \midrule
 &
  SAM &
  0.75 ± 0.92 &
  0.85 ± 0.6 &
  1.55 ± 1.29 &
  3.93 ± 2.73 &
  0.3 ± 0.42 &
  0.08 ± 0.29 &
  1.88 ± 1.59 \\ \cmidrule(l){2-9} 
\multirow{-2}{*}{Facial Asymmetry} &
  DLV3 &
  \textbf{0.14 ± 0.13} &
  \textbf{0.19 ± 0.07} &
  \textbf{0.60 ± 0.36} &
  \textbf{1.30 ± 0.71} &
  \textbf{0.00 ± 0.01} &
  \textbf{0.00 ± 0.00} &
  \textbf{0.63 ± 0.34} \\ \midrule
 &
  SAM &
  0.63 ± 0.44 &
  0.65 ± 0.48 &
  2.38 ± 2.54 &
  4.15 ± 3.66 &
  0.23 ± 0.38 &
  0.01 ± 0.05 &
  1.78 ± 1.41 \\ \cmidrule(l){2-9} 
\multirow{-2}{*}{Syndrome} &
  DLV3 &
  \textbf{0.41 ± 0.15} &
  \textbf{0.29 ± 0.19} &
  \textbf{0.91 ± 0.63} &
  \textbf{1.40 ± 0.90} &
  \textbf{0.05 ± 0.11} &
  \textbf{0.00 ± 0.00} &
  \textbf{0.69 ± 0.30} \\ \midrule
 &
  SAM &
  0.5 ± 0.32 &
  0.69 ± 0.73 &
  2.2 ± 2.63 &
  4.46 ± 4.66 &
  0.28 ± 0.62 &
  0.03 ± 0.18 &
  1.89 ± 1.85 \\ \cmidrule(l){2-9} 
\multirow{-2}{*}{Unknown/Other Craniofacial} &
  DLV3 &
  \textbf{0.37 ± 0.15} &
  \textbf{0.27 ± 0.15} &
  \textbf{0.74 ± 0.53} &
  \textbf{1.53 ± 1.07} &
  \textbf{0.01 ± 0.04} &
  \textbf{0.00 ± 0.00} &
  \textbf{0.55 ± 0.38} \\ \bottomrule
\end{tabular}%
}
\caption{Mean Absolute Error (MAE) of all models on all datasets used in this study for eye measurements. MAE was calculated according to Equation \ref{mae}, and is reported as +/- the standard deviation. Bold indicates the lowest MAE for each measurement for each model. Bilateral distances were averaged. Red denotes the case where SAM outperformed the DLV3 model. Abbreviations can be interpreted as follows: MRD-Margin to Reflex Distance, ICD-inner canthal distance, IPD-interpupillary distance, OCD- outer canthal distance, ISS- inferior scleral show, SSS- superior scleral show, Horiz Fissure- horizontal palpebral fissure. }
\label{tab:mae}
\end{table}

\begin{table}[]
\centering
\resizebox{\columnwidth}{!}{%
\begin{tabular}{@{}|c|c|cccc|cccc|cc|@{}}
\toprule
\cellcolor[HTML]{FFFFFF} &
   &
  \multicolumn{4}{c|}{\textbf{Right}} &
  \multicolumn{4}{c|}{\textbf{Left}} &
  \multicolumn{2}{c|}{\textbf{Bilateral}} \\ \midrule
 &
   &
  \multicolumn{1}{c|}{\textbf{MRD 1}} &
  \multicolumn{1}{c|}{\textbf{MRD 2}} &
  \multicolumn{1}{c|}{\textbf{Lat Canthal}} &
  \textbf{Med Canthal} &
  \multicolumn{1}{c|}{\textbf{MRD 1}} &
  \multicolumn{1}{c|}{\textbf{MRD 2}} &
  \multicolumn{1}{c|}{\textbf{Lat Canthal}} &
  \textbf{Med Canthal} &
  \multicolumn{1}{c|}{\textbf{ICD}} &
  \textbf{OCD} \\ \midrule
 &
  \textbf{\% Dataset} &
  \multicolumn{2}{c|}{0.85} &
  \multicolumn{2}{c|}{0.85} &
  \multicolumn{2}{c|}{0.86} &
  \multicolumn{2}{c|}{0.86} &
  \multicolumn{2}{c|}{0.82} \\ \cmidrule(l){2-12} 
 &
  \textbf{periorbitAI} &
  \multicolumn{1}{c|}{1.21 ± 0.87} &
  \multicolumn{1}{c|}{0.54 ± 0.52} &
  \multicolumn{1}{c|}{0.94 ± 0.75} &
  0.71 ± 0.60 &
  \multicolumn{1}{c|}{1.08 ± 0.88} &
  \multicolumn{1}{c|}{0.50 ± 0.47} &
  \multicolumn{1}{c|}{0.79 ± 0.63} &
  0.73 ± 0.60 &
  \multicolumn{1}{c|}{3.82 ± 3.99} &
  4.99 ± 4.09 \\ \cmidrule(l){2-12} 
 &
  \textbf{SAM} &
  \multicolumn{1}{c|}{0.59 ± 0.65} &
  \multicolumn{1}{c|}{\textbf{0.45 ± 0.44}} &
  \multicolumn{1}{c|}{0.67 ± 0.66} &
  \textbf{0.69 ± 0.62} &
  \multicolumn{1}{c|}{0.61 ± 0.74} &
  \multicolumn{1}{c|}{\textbf{0.43 ± 0.58}} &
  \multicolumn{1}{c|}{0.63 ± 0.62} &
  \textbf{0.66 ± 0.63} &
  \multicolumn{1}{c|}{2.44 ± 2.63} &
  2.86 ± 3.18 \\ \cmidrule(l){2-12} 
\multirow{-4}{*}{\textbf{Healthy}} &
  \textbf{DLV3} &
  \multicolumn{1}{c|}{\textbf{0.35 ± 0.32}} &
  \multicolumn{1}{c|}{0.68 ± 0.40} &
  \multicolumn{1}{c|}{\textbf{0.51 ± 0.43}} &
  0.72 ± 0.55 &
  \multicolumn{1}{c|}{\textbf{0.30 ± 0.28}} &
  \multicolumn{1}{c|}{0.61 ± 0.34} &
  \multicolumn{1}{c|}{\textbf{0.50 ± 0.40}} &
  0.67 ± 0.51 &
  \multicolumn{1}{c|}{3.60 ± 1.80} &
  \textbf{1.37 ± 1.79} \\ \midrule
 &
  \textbf{\% Dataset} &
  \multicolumn{2}{c|}{0.6} &
  \multicolumn{2}{c|}{0.58} &
  \multicolumn{2}{c|}{0.64} &
  \multicolumn{2}{c|}{0.63} &
  \multicolumn{2}{c|}{0.52} \\ \cmidrule(l){2-12} 
 &
  \textbf{periorbitAI} &
  \multicolumn{1}{c|}{0.88 ± 1.23} &
  \multicolumn{1}{c|}{0.93 ± 0.91} &
  \multicolumn{1}{c|}{1.15 ± 1.18} &
  1.00 ± 0.89 &
  \multicolumn{1}{c|}{1.10 ± 1.62} &
  \multicolumn{1}{c|}{0.90 ± 1.09} &
  \multicolumn{1}{c|}{1.29 ± 1.35} &
  1.05 ± 1.13 &
  \multicolumn{1}{c|}{7.12 ± 5.30} &
  \textbf{{\color{red}3.07 ± 2.89}} \\ \cmidrule(l){2-12} 
 &
  \textbf{SAM} &
  \multicolumn{1}{c|}{0.77 ± 0.84} &
  \multicolumn{1}{c|}{0.84 ± 0.96} &
  \multicolumn{1}{c|}{0.84 ± 0.61} &
  \textbf{0.62 ± 0.59} &
  \multicolumn{1}{c|}{0.86 ± 1.07} &
  \multicolumn{1}{c|}{\textbf{1.38 ± 1.81}} &
  \multicolumn{1}{c|}{1.01 ± 0.74} &
  0.84 ± 0.81 &
  \multicolumn{1}{c|}{4.06 ± 3.85} &
  4.85 ± 3.13 \\ \cmidrule(l){2-12} 
\multirow{-4}{*}{\textbf{TED}} &
  \textbf{DLV3} &
  \multicolumn{1}{c|}{\textbf{0.57 ± 0.59}} &
  \multicolumn{1}{c|}{\textbf{0.54 ± 0.60}} &
  \multicolumn{1}{c|}{\textbf{0.67 ± 0.64}} &
  0.73 ± 0.64 &
  \multicolumn{1}{c|}{\textbf{0.57 ± 0.76}} &
  \multicolumn{1}{c|}{0.53 ± 0.42} &
  \multicolumn{1}{c|}{\textbf{0.81 ± 0.70}} &
  \textbf{0.59 ± 0.54} &
  \multicolumn{1}{c|}{\textbf{2.29 ± 2.43}} &
  3.76 ± 2.36 \\ \midrule
 &
  \textbf{\% Dataset} &
  \multicolumn{2}{c|}{0.85} &
  \multicolumn{2}{c|}{0.69} &
  \multicolumn{2}{c|}{0.81} &
  \multicolumn{2}{c|}{0.66} &
  \multicolumn{2}{c|}{0.69} \\ \cmidrule(l){2-12} 
 &
  \textbf{periorbitAI} &
  \multicolumn{1}{c|}{1.69  ± 1.41} &
  \multicolumn{1}{c|}{1.58  ± 1.66} &
  \multicolumn{1}{c|}{1.21  ± 0.72} &
  1.53  ± 1.24 &
  \multicolumn{1}{c|}{1.82    ± 1.34} &
  \multicolumn{1}{c|}{1.50  ± 1.45} &
  \multicolumn{1}{c|}{1.23  ± 1.31} &
  1.83  ± 1.66 &
  \multicolumn{1}{c|}{14.23  ± 12.17} &
  26.44  ± 32.54 \\ \cmidrule(l){2-12} 
 &
  \textbf{SAM} &
  \multicolumn{1}{c|}{0.59  ± 0.53} &
  \multicolumn{1}{c|}{0.63  ± 0.62} &
  \multicolumn{1}{c|}{0.82  ± 0.73} &
  \textbf{0.57  ± 0.56} &
  \multicolumn{1}{c|}{0.62    ± 0.55} &
  \multicolumn{1}{c|}{0.63  ± 0.61} &
  \multicolumn{1}{c|}{1.19  ± 1.23} &
  0.57  ± 0.68 &
  \multicolumn{1}{c|}{2.04  ± 1.71} &
  3.93  ± 2.97 \\ \cmidrule(l){2-12} 
\multirow{-4}{*}{\textbf{Syndrome}} &
  \textbf{DLV3} &
  \multicolumn{1}{c|}{\textbf{0.52  ± 0.42}} &
  \multicolumn{1}{c|}{\textbf{0.62  ± 0.53}} &
  \multicolumn{1}{c|}{\textbf{0.75  ± 0.64}} &
  0.75  ± 0.55 &
  \multicolumn{1}{c|}{\textbf{0.41    ± 0.38}} &
  \multicolumn{1}{c|}{\textbf{0.42  ± 0.43}} &
  \multicolumn{1}{c|}{\textbf{0.59  ± 0.54}} &
  \textbf{0.54  ± 0.52} &
  \multicolumn{1}{c|}{\textbf{1.11  ± 1.07}} &
  \textbf{3.42  ± 2.71} \\ \midrule
 &
  \textbf{\% Dataset} &
  \multicolumn{2}{c|}{1} &
  \multicolumn{2}{c|}{1} &
  \multicolumn{2}{c|}{0.82} &
  \multicolumn{2}{c|}{0.82} &
  \multicolumn{2}{c|}{0.82} \\ \cmidrule(l){2-12} 
 &
  \textbf{periorbitAI} &
  \multicolumn{1}{c|}{1.13  ± 1.18} &
  \multicolumn{1}{c|}{\textbf{{\color{red}0.26  ± 0.27}}} &
  \multicolumn{1}{c|}{0.58  ± 0.63} &
  1.14  ± 0.80 &
  \multicolumn{1}{c|}{1.39    ± 1.37} &
  \multicolumn{1}{c|}{2.10  ± 2.45} &
  \multicolumn{1}{c|}{0.69  ± 0.29} &
  1.51  ± 1.27 &
  \multicolumn{1}{c|}{13.48  ± 14.89} &
  30.07  ± 37.28 \\ \cmidrule(l){2-12} 
 &
  \textbf{SAM} &
  \multicolumn{1}{c|}{0.60  ± 0.57} &
  \multicolumn{1}{c|}{0.70  ± 0.75} &
  \multicolumn{1}{c|}{0.77  ± 0.67} &
  0.77  ± 0.55 &
  \multicolumn{1}{c|}{0.93    ± 1.39} &
  \multicolumn{1}{c|}{1.05  ± 0.95} &
  \multicolumn{1}{c|}{0.82  ± 0.71} &
  0.68  ± 0.61 &
  \multicolumn{1}{c|}{1.47  ± 1.32} &
  4.01  ± 2.84 \\ \cmidrule(l){2-12} 
\multirow{-4}{*}{\textbf{Facial Asymmetry}} &
  \textbf{DLV3} &
  \multicolumn{1}{c|}{\textbf{0.49  ± 0.44}} &
  \multicolumn{1}{c|}{0.56  ± 0.45} &
  \multicolumn{1}{c|}{\textbf{0.56  ± 0.46}} &
  \textbf{0.72  ± 0.51} &
  \multicolumn{1}{c|}{\textbf{0.52    ± 0.26}} &
  \multicolumn{1}{c|}{\textbf{0.36  ± 0.31}} &
  \multicolumn{1}{c|}{\textbf{0.53  ± 0.41}} &
  \textbf{0.64  ± 0.56} &
  \multicolumn{1}{c|}{\textbf{1.14  ± 1.07}} &
  \textbf{3.49  ± 1.45} \\ \midrule
 &
  \textbf{\% Dataset} &
  \multicolumn{2}{c|}{0.88} &
  \multicolumn{2}{c|}{0.88} &
  \multicolumn{2}{c|}{1} &
  \multicolumn{2}{c|}{1} &
  \multicolumn{2}{c|}{0.88} \\ \cmidrule(l){2-12} 
 &
  \textbf{periorbitAI} &
  \multicolumn{1}{c|}{1.56  ± 1.09} &
  \multicolumn{1}{c|}{0.95  ± 0.75} &
  \multicolumn{1}{c|}{0.63  ± 0.79} &
  1.36  ± 0.88 &
  \multicolumn{1}{c|}{4.14    ± 3.15} &
  \multicolumn{1}{c|}{0.99  ± 1.59} &
  \multicolumn{1}{c|}{1.89  ± 2.13} &
  1.40  ± 1.83 &
  \multicolumn{1}{c|}{8.15  ± 11.45} &
  16.13  ± 28.12 \\ \cmidrule(l){2-12} 
 &
  \textbf{SAM} &
  \multicolumn{1}{c|}{0.51  ± 0.51} &
  \multicolumn{1}{c|}{0.56  ± 0.43} &
  \multicolumn{1}{c|}{0.53  ± 0.31} &
  0.84  ± 0.54 &
  \multicolumn{1}{c|}{0.52    ± 0.21} &
  \multicolumn{1}{c|}{0.52  ± 0.40} &
  \multicolumn{1}{c|}{1.20  ± 0.99} &
  \textbf{0.53  ± 0.33} &
  \multicolumn{1}{c|}{1.83  ± 1.48} &
  \textbf{3.28  ± 2.39} \\ \cmidrule(l){2-12} 
\multirow{-4}{*}{\textbf{Craniosynostosis}} &
  \textbf{DLV3} &
  \multicolumn{1}{c|}{\textbf{0.47  ± 0.52}} &
  \multicolumn{1}{c|}{\textbf{0.46  ± 0.33}} &
  \multicolumn{1}{c|}{\textbf{0.26  ± 0.28}} &
  \textbf{0.76  ± 0.45} &
  \multicolumn{1}{c|}{\textbf{0.35    ± 0.37}} &
  \multicolumn{1}{c|}{\textbf{0.45  ± 0.34}} &
  \multicolumn{1}{c|}{\textbf{0.81  ± 0.78}} &
  0.62  ± 0.65 &
  \multicolumn{1}{c|}{\textbf{1.46  ± 1.54}} &
  3.34  ± 2.21 \\ \midrule
 &
  \textbf{\% Dataset} &
  \multicolumn{2}{c|}{0.9} &
  \multicolumn{2}{c|}{0.8} &
  \multicolumn{2}{c|}{0.9} &
  \multicolumn{2}{c|}{0.7} &
  \multicolumn{2}{c|}{0.8} \\ \cmidrule(l){2-12} 
 &
  \textbf{periorbitAI} &
  \multicolumn{1}{c|}{3.13  ± 2.13} &
  \multicolumn{1}{c|}{1.02  ± 0.91} &
  \multicolumn{1}{c|}{0.85  ± 0.70} &
  0.66  ± 0.33 &
  \multicolumn{1}{c|}{2.37    ± 2.19} &
  \multicolumn{1}{c|}{1.20  ± 1.44} &
  \multicolumn{1}{c|}{1.37  ± 0.74} &
  0.50  ± 0.38 &
  \multicolumn{1}{c|}{17.24  ± 16.18} &
  36.28  ± 38.49 \\ \cmidrule(l){2-12} 
 &
  \textbf{SAM} &
  \multicolumn{1}{c|}{0.34  ± 0.30} &
  \multicolumn{1}{c|}{0.65  ± 0.67} &
  \multicolumn{1}{c|}{\textbf{0.54  ± 0.34}} &
  0.49  ± 0.47 &
  \multicolumn{1}{c|}{0.63    ± 0.65} &
  \multicolumn{1}{c|}{1.10  ± 1.27} &
  \multicolumn{1}{c|}{0.73  ± 0.59} &
  \textbf{0.44  ± 0.27} &
  \multicolumn{1}{c|}{2.03  ± 1.62} &
  \textbf{3.24  ± 1.97} \\ \cmidrule(l){2-12} 
\multirow{-4}{*}{\textbf{Cleft}} &
  \textbf{DLV3} &
  \multicolumn{1}{c|}{\textbf{0.50  ± 0.43}} &
  \multicolumn{1}{c|}{\textbf{0.59  ± 0.44}} &
  \multicolumn{1}{c|}{0.55  ± 0.27} &
  \textbf{0.46  ± 0.26} &
  \multicolumn{1}{c|}{\textbf{0.46    ± 0.28}} &
  \multicolumn{1}{c|}{\textbf{0.60  ± 0.44}} &
  \multicolumn{1}{c|}{\textbf{0.69  ± 0.51}} &
  0.46  ± 0.27 &
  \multicolumn{1}{c|}{\textbf{0.63  ± 0.63}} &
  3.40  ± 2.68 \\ \midrule
 &
  \textbf{\% Dataset} &
  \multicolumn{2}{c|}{0.95} &
  \multicolumn{2}{c|}{0.84} &
  \multicolumn{2}{c|}{0.84} &
  \multicolumn{2}{c|}{0.77} &
  \multicolumn{2}{c|}{0.79} \\ \cmidrule(l){2-12} 
 &
  \textbf{periorbitAI} &
  \multicolumn{1}{c|}{1.50 ± 1.37} &
  \multicolumn{1}{c|}{1.23 ± 1.19} &
  \multicolumn{1}{c|}{1.33 ± 0.81} &
  1.04 ± 0.78 &
  \multicolumn{1}{c|}{1.52 ± 1.35} &
  \multicolumn{1}{c|}{1.24 ± 1.64} &
  \multicolumn{1}{c|}{1.11 ± 0.87} &
  1.14 ± 0.74 &
  \multicolumn{1}{c|}{12.76 ± 12.79} &
  23.42 ± 33.56 \\ \cmidrule(l){2-12} 
 &
  \textbf{SAM} &
  \multicolumn{1}{c|}{\textbf{0.47 ± 0.32}} &
  \multicolumn{1}{c|}{0.70 ± 0.72} &
  \multicolumn{1}{c|}{0.78 ± 0.72} &
  \textbf{0.54 ± 0.35} &
  \multicolumn{1}{c|}{0.52 ± 0.64} &
  \multicolumn{1}{c|}{0.68 ± 1.16} &
  \multicolumn{1}{c|}{0.65 ± 0.66} &
  \textbf{0.52 ± 0.66} &
  \multicolumn{1}{c|}{1.82 ± 1.80} &
  4.25 ± 3.45 \\ \cmidrule(l){2-12} 
\multirow{-4}{*}{\textbf{Unknown/Other Craniofacial}} &
  \textbf{DLV3} &
  \multicolumn{1}{c|}{0.48 ± 0.45} &
  \multicolumn{1}{c|}{\textbf{0.52 ± 0.45}} &
  \multicolumn{1}{c|}{\textbf{0.55 ± 0.41}} &
  0.63 ± 0.56 &
  \multicolumn{1}{c|}{\textbf{0.39 ± 0.26}} &
  \multicolumn{1}{c|}{\textbf{0.49 ± 0.46}} &
  \multicolumn{1}{c|}{\textbf{0.51 ± 0.47}} &
  0.62 ± 0.48 &
  \multicolumn{1}{c|}{\textbf{1.13 ± 1.26}} &
  \textbf{3.44 ± 2.14} \\ \bottomrule
\end{tabular}}
\caption{Comparison of MAE (Equation \ref{mae}) of our models to PeriorbitAI for eye measurements.
For all measurements, for both our models and PeriorbitAI, MAE was computed using only
images successfully analyzed by PeriorbitAI. ‘\% Dataset’ denotes the percentage of the original
dataset for each measurement successfully processed by PeriorbitAI. Bold denotes lowest MAE
of each measurement for each dataset. Red denotes cases where PeriorbitAI outperformed both of our models. Abbreviations can be interpreted as follows: MRD-Margin
to Reflex Distance, ICD-inner canthal distance, IPD-interpupillary distance, OCD- outer canthal
distance, ISS- inferior scleral show, SSS- superior scleral show, Vert Fissure- vertical palpebral
fissure, Horiz Fissure- horizontal palpebral fissure, Med Canthal- medial canthal height, Lat
Canthal- lateral canthal height.}
\label{tab:pai}
\end{table}

\begin{figure} [!hb]
    \centering
    \includegraphics[width=1\linewidth]{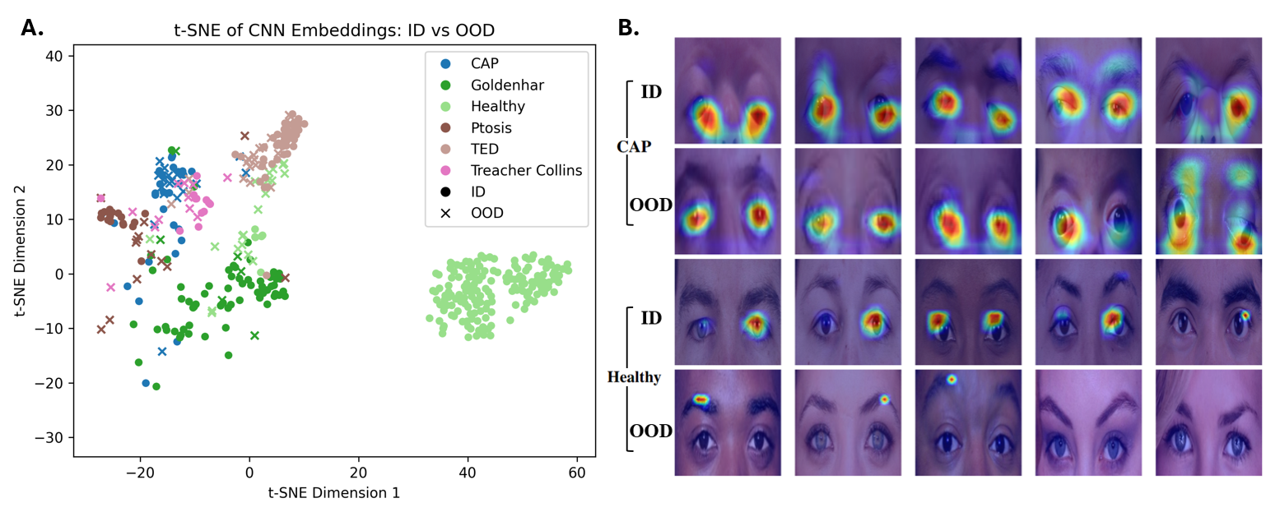}
    \caption{t-SNE plot of embeddings of ID test and OOD from finetuned ResNet-18. X's represent OOD and O's represent ID train samples. B) Grad-CAM visualizations of a CNN classifier trained on ID data for randomly sampled images of CAP (crouzon-apert-pfeiffer) and Healthy classes from ID and OOD datasets. CAP had the lowest difference in Wasserstein distance between OOD-ID train and ID test-ID train, and Healthy had the highest}
    \label{fig:tsne-gradcam}
\end{figure}

\subsection{Classification Experiments}

We evaluated disease classification performance using both CNN-based image models and tabular models trained on periorbital distances. This allowed us to assess the generalizability and feature robustness of each modeling approach across distribution shifts. All models were trained exclusively on ID data and evaluated on both ID holdout sets and OOD datasets collected from the web (see graphical schematic in Figure \ref{fig:ml_schem}).

\subsubsection{Classification Performance}

On the ID test set, CNN-based models achieved the highest accuracy (78\%), with XGBoost models trained on periorbital distances closely behind (77\%). Lasso regression lagged slightly (66\%), but all three models achieved similar AUROC values around 0.96–0.97, indicating strong discriminative performance (Table \ref{tab:ml}).

Fusion models, which combined CNN embeddings with periorbital distances, yielded the best overall performance on ID data (accuracy 80\%, F1 score 0.80, AUROC 0.96).

On OOD data, CNN performance dropped sharply (accuracy 14\%, AUROC 0.59), consistent with the effects of domain shift \cite{rashidisabet2023validating}. In contrast, Lasso and XGBoost models using periorbital distances showed strong generalization, achieving AUROCs of 0.93 and 0.91, respectively. XGBoost also achieved the highest OOD precision (PPV 0.87), while Lasso achieved the highest OOD accuracy (68\%).

Fusion models demonstrated intermediate performance on OOD data (accuracy 67\%, AUROC 0.79), outperforming CNNs but underperforming relative to the shallow models.

\subsubsection{Model Behavior and Domain Shift}

CNN embeddings for ID and OOD samples formed distinct clusters per class following dimensionality reduction t-SNE space (Figure \ref{fig:tsne-gradcam}A), confirming distributional drift. This was further supported by Wasserstein distance analyses, where every OOD class was more distant from the ID training distribution than its ID test counterpart (Supplemental Table \ref{tab:wasserstein}).

To understand how this shift affected CNN decision-making, we used Grad-CAM to visualize regions of model attention for the classes with the largest and smallest Wasserstein distance deltas relative to the ID train distribution(\cite{selvaraju2020grad}). While attention in the low-shift class (CAP) remained centered on periorbital features, the high-shift class (Healthy Adult) showed diffuse and inconsistent attention, suggesting less reliable feature attribution.(Figure \ref{fig:tsne-gradcam}B). 

\subsubsection{Effect of Pediatric Status and Importances}

Pediatric status was among the most predictive features in both Lasso and XGBoost models. To assess its influence on model behavior, we conducted ablation experiments by removing this feature. This resulted in a measurable drop in ID classification performance (Supplemental Figure \ref{tab:ped-ablation}), highlighting its importance as a discriminative attribute in our dataset.

After retraining without pediatric status, feature importance shifted toward anatomical features such as interpupillary distance, intercanthal distance, and asymmetry metrics capturing left–right differences in periorbital structure. These features were consistently ranked among the top predictors across disease groups (Supplemental Figure \ref{tab:feat-import}).

\begin{table}[]
\centering
\resizebox{\columnwidth}{!}{%
\begin{tabular}{@{}|c|c|c|c|c|c|c|@{}}
\toprule
\cellcolor[HTML]{FFFFFF}\textbf{Dataset} & Classification & Accuracy      & Recall        & PPV           & F1 Score      & AUROC              \\ \midrule
                      & XGBoost    & 0.77          & 0.75          & 0.77          & 0.76          & \textbf{0.97 {[}.96-.98{]}} \\ \cmidrule(l){2-7} 
                      & Lasso  & 0.66          & 0.63          & 0.67          & 0.65          & 0.96 {[}.95-.96{]}          \\ \cmidrule(l){2-7} 
                      & CNN    & 0.78          & 0.78          & 0.80          & 0.78          & 0.96 {[}.95-.97{]}          \\ \cmidrule(l){2-7} 
\multirow{-4}{*}{ID}                     & Fusion         & \textbf{0.80} & \textbf{0.80} & \textbf{0.81} & \textbf{0.80} & 0.96 {[}.95-.97{]} \\ \midrule
                      & XGBoost    & 0.63          & 0.58          & \textbf{0.87} & \textbf{0.69} & \textbf{0.91 {[}.86-.95{]}} \\ \cmidrule(l){2-7} 
                      & Lasso  & \textbf{0.68} & 0.64          & 0.71          & 0.67          & 0.93 {[}.89-.96{]}          \\ \cmidrule(l){2-7} 
                      & CNN    & 0.14          & 0.10          & 0.36          & 0.16          & 0.59 {[}.52-.64{]}          \\ \cmidrule(l){2-7} 
\multirow{-4}{*}{OOD} & Fusion & 0.67          & \textbf{0.67} & 0.61          & 0.62          & .79 {[}.70-.88{]}           \\ \bottomrule
\end{tabular}%
}
\caption{Classification results. XGBoost and Lasso denotes an XG-Boost, or Lasso model trained using  periorbital distances. ID and OOD denote in and out of distribution datasets, respectively. Fusion denotes a multi-layer perceptron trained using periorbital features as well as CNN features.}
\label{tab:ml}
\end{table}

\section{Discussion}\label{disc}
Prior efforts have focused on automating periorbital distance measurement using deep learning. Shao et al. developed a segmentation pipeline for assessing eyelid morphology and predicting MRD 1 and MRD 2 in thyroid eye disease (TED), while Chen et al. proposed a smartphone-based model to estimate MRD values and levator function \cite{chen_smartphone-based_2021, shao_deep_2023}. Rana et al. analyzed periorbital distances in healthy individuals across ethnicities \cite{rana_artificial_2024}, and Van Brummen et al. introduced PeriorbitAI, a multi-measurement model evaluated across several conditions but limited to standardized imaging and a small test set (n=41) \cite{van_brummen_periorbitai_2021}. To the best of our knowledge, no prior study has explored the use of periorbital measurements for automated disease classification, particularly in the context of domain shift and real-world generalizability.

\begin{figure}
    \centering
    \includegraphics[width=1\linewidth]{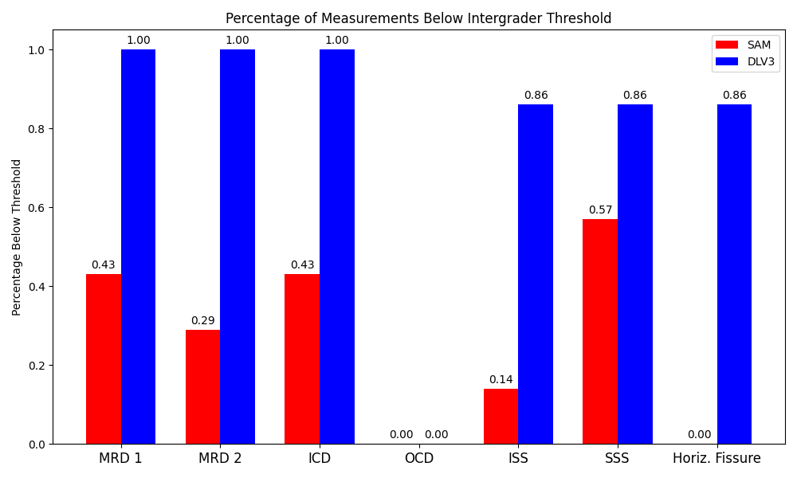}
    \caption{Percentage of eye measurements below intergrader threshold across all datrasets for SAM and DLV3 segmentation backbones.}
    \label{fig:threshold}
\end{figure}

Our work builds on this foundation by presenting a segmentation pipeline that achieves state-of-the-art performance for periorbital distance prediction across multiple disease classes. By training DeepLabV3 on a domain-specific dataset of healthy eyes annotated for periorbital anatomy, we were able to achieve robust segmentation across diverse, non-standardized clinical images. Compared to the current benchmark (PeriorbitAI), our model yielded lower mean absolute errors on nearly all measurements and disease groups, including images captured on smartphones (Table \ref{tab:pai}). Across most measurements, our model achieved accuracy within the range of intergrader variability (Supplemental Figure \ref{fig:threshold}, Table \ref{tab:mae}), establishing it as the most comprehensive and accurate method to date for automated periorbital distance quantification (Table \ref{tab:pai}).

We also compared DeepLabV3 to the Segment Anything Model (SAM), a foundational model trained on a wide range of image domains. Despite SAM’s generalization capacity, it underperformed relative to our best model by an average of 0.8 mm across measurements. These findings suggest that for tasks requiring anatomical precision, such as distance-based measurement, tailored segmentation models trained on task-relevant annotations remain superior to large, general-purpose models.

Beyond segmentation, we demonstrate that these predicted distances can serve as robust features for AI facilitated disease classification. On ID data, shallow models using periorbital distances performed comparably to image-based CNN classifiers (77\% vs. 78\% accuracy; Table \ref{tab:ml}). However, under OOD conditions—where CNN performance dropped dramatically (14\% accuracy, AUROC 0.59), periorbital distance-based models generalized substantially better (XGBoost accuracy: 63\%, AUROC: 0.91) despite being solely trained on ID data. These results were supported by t-SNE visualizations, Grad-CAM saliency maps, and Wasserstein distance metrics, all of which highlighted the CNN’s sensitivity to domain shift (Figure \ref{fig:tsne-gradcam}, Supplemental Table \ref{tab:wasserstein}). This sensitivity to domain shift has been reported in the literature by many previous studies (\cite{rashidisabet2023validating,stacke2019closer,pooch2019can,moreno2012unifying}).

The fusion model, which combined CNN-derived image embeddings with segmentation-based periorbital distances, achieved the highest performance on ID data (80\% accuracy, AUROC 0.96; Table \ref{tab:ml}). However, its performance declined on OOD data, underperforming relative to models using periorbital distances alone (67\% accuracy, AUROC 0.79). This was an expected finding, given that CNN features degraded substantially under domain shift and contributed noisy or non-informative representations. These results suggest that in its current form, the fusion model may inherit the domain sensitivity of the CNN backbone. However, this architecture provides a promising framework for future development, and future advances in multimodal learning architectures could help overcome this limitation.. 

The use of periorbital distances as classification features offers practical clinical advantages. These measurements are anatomically interpretable and can be extracted from real-world images without standardized lighting, pose, or background. As such, segmentation-derived features may enable explainable and robust disease classification in diverse clinical environments, including resource-limited settings where specialist expertise is unavailable. For example, one study found that 89\% of U.S. counties do not have access to an oculoplastic surgeon (\cite{hussey2022oculofacial}). Incorporating lightweight, portable diagnostic tools based on segmentation and distance prediction may help extend specialist-level diagnostic capabilities to under-served populations (\cite{van_brummen_periorbitai_2021}).

Future work in this domain could proceed along several promising paths. First, classification performance may be improved through the development of more advanced fusion architectures. Attention-based or modality-aware models that dynamically weight contributions from image and distance features could better leverage complementary information while mitigating the impact of degraded image embeddings in OOD scenarios \cite{du2024tip}. Second, segmentation accuracy could benefit from leveraging large-scale unlabeled datasets. Semi-supervised and self-supervised learning strategies—particularly those designed for medical imaging—offer a path toward improved generalization without requiring extensive manual annotation \cite{yang2022st++, yang2025unimatch}. Finally, the deployment of portable, privacy-preserving platforms for periorbital measurement could support real-time analysis in diverse clinical and field environments, expanding access to automated diagnostics in oculoplastic and craniofacial care. Together, these directions may help realize the potential of segmentation-driven AI systems for robust, accessible craniofacial and oculoplastic care.

\subsection*{Limitations}

This study has several limitations. First, the OOD dataset was relatively small, which limits our ability to fully assess generalizability. Although periorbital distances demonstrated robustness under distribution shift, broader validation on external datasets is warranted. Second, pediatric status emerged as a highly predictive feature in classification models, which may reflect true anatomical differences but also raises concerns about overfitting to age-related cues. While we performed ablation experiments to assess its influence, future work should examine how age-specific models or age-matched cohorts affect performance. Third, the accuracy of distance prediction depends on consistent iris segmentation, which can be challenging in cases of low image quality or occlusion. Lastly, disease labels were assigned based on clinical diagnosis rather than genetic confirmation, which may introduce label variability.

\section{Dataset and Code Availability}

We have made our public OOD dataset available for download \href{https://drive.google.com/drive/folders/1aPdEuIHHsBLZH3pSsQAQdjY1Z06-rHDN?usp=sharing} {here}. All code used for periorbital distance prediction is publicly available and accessible via the standard Python \href{https://github.com/monkeygobah/periorbital_package}{package installer}.

\bibliographystyle{unsrt}
\bibliography{main}

\section{Supplemental Figures}\label{figs}
\newpage

\begin{figure} [t]
    \centering
    \includegraphics[width=.75\linewidth]{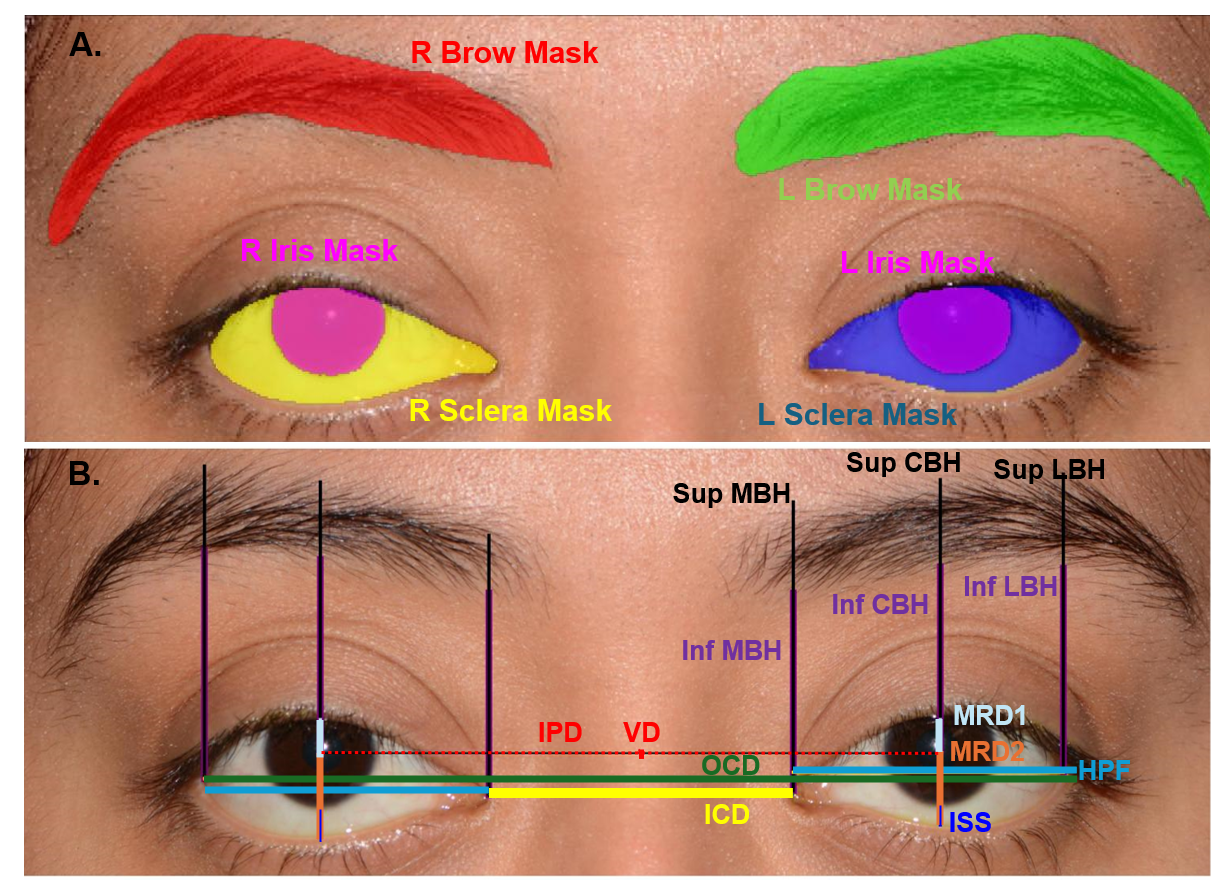}
    \caption{Representative example of ground truth segmentation masks and periorbital distance
prediction. A): Ground truth masks of the anatomical regions used for evaluating segmentation results and
deriving ground truth distance measurements on the face. B). Distance measurements calculated
from (A). Pixels were converted to mm using 11.71 mm as the standard diameter for the iris.
Scleral area was calculated by taking the ratio of sclera mask to the iris mask, and 4th degree
polynomials were fit to the superior and inferior scleral margin. Abbreviations are as follows:
VD- Vertical Dystopia, IPD-Inner Pupil Distance, OCD-Outer Canthal Distance, ICD-Inner
Canthal Distance, HPF-Horizontal Palpebral Fissure, MRD-Margin to Reflex Distance, ISS-
Inferior Scleral Show. Other measurements not shown are VPF-sum of MRD 1 and 2, canthal
height-distance between inner pupillary line and medial/lateral canthus, canthal tilt, and scleral
area.}
    \label{fig:fig_1}
\end{figure}

% \begin{figure}[H]
%     \centering
%     \includegraphics[width=1\linewidth]{dice.png}
%     \caption{{\color{blue}A) Dice score for SAM iris segmentation on all three datasets. Red line
% indicates the median, and the lines indicate the upper and lower quartiles. B-D) Dice scores for sclera and E-G) brow segmentation on all three datasets using SAM, and DeepLabV3 when trained on increasingly large datasets. The red line indicates average SAM dice score, red shade indicates the standard deviation across the entire dataset. Green (DeepLabV3) lines and bars denote the average and standard deviation segmentation at all of the training sizes evaluated.}}
%     \label{fig:dice}
% \end{figure}

\begin{figure}[H]
    \centering
    \includegraphics[width=1.\linewidth]{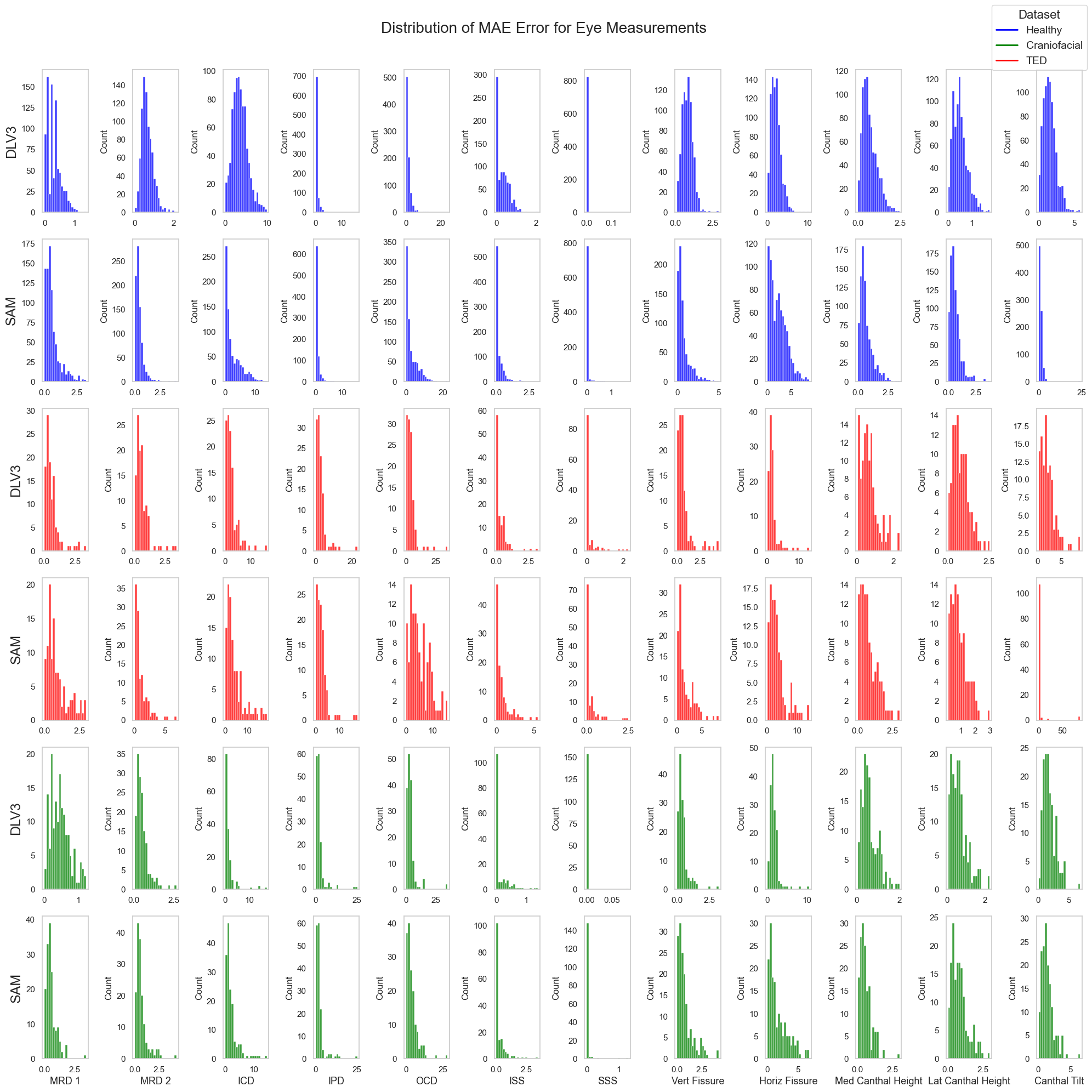}
    \caption{Histograms of MAE for all eye measurements using all models for all datasets. Bilateral distances were averaged. Scale is in mm following conversion using the iris as a diameter.}
    \label{fig:eye_histos}
\end{figure}

\newpage

\section{Supplemental Tables}\label{tabs}
\newpage

% Please add the following required packages to your document preamble:
% \usepackage{booktabs}
% \usepackage{graphicx}
% \usepackage[table,xcdraw]{xcolor}
% Beamer presentation requires \usepackage{colortbl} instead of \usepackage[table,xcdraw]{xcolor}
\begin{table}[]
\centering
\resizebox{\columnwidth}{!}{%
\begin{tabular}{@{}|c|c|cc|cc|@{}}
\toprule
\cellcolor[HTML]{FFFFFF}\textbf{Disease} &
  \textbf{Big Bucket} &
  \multicolumn{2}{c|}{\textbf{Count}} &
  \multicolumn{2}{c|}{\textbf{Big Bucket}} \\ \midrule
goldenhar syndrome &
  Syndrome &
  \multicolumn{1}{c|}{26} &
  craniosynostosis - sagittal &
  \multicolumn{1}{c|}{Facial Asymmetry} &
  1 \\ \midrule
treacher collins syndrome &
  Syndrome &
  \multicolumn{1}{c|}{7} &
  craniosynostosis - pansynostosis &
  \multicolumn{1}{c|}{Facial Asymmetry} &
  1 \\ \midrule
pierre robin sequence &
  Syndrome &
  \multicolumn{1}{c|}{4} &
  craniosynostosis - coronal - unilateral &
  \multicolumn{1}{c|}{Craniosynostosis} &
  3 \\ \midrule
nager syndrome &
  Syndrome &
  \multicolumn{1}{c|}{3} &
  craniosynostosis - coronal - bilateral &
  \multicolumn{1}{c|}{Craniosynostosis} &
  1 \\ \midrule
parry romberg disease &
  Syndrome &
  \multicolumn{1}{c|}{3} &
  craniosynostosis - metopic &
  \multicolumn{1}{c|}{Craniosynostosis} &
  1 \\ \midrule
stickler syndrome &
  Syndrome &
  \multicolumn{1}{c|}{2} &
  craniosynostosis, not classified &
  \multicolumn{1}{c|}{Craniosynostosis} &
  3 \\ \midrule
noonan syndrome &
  Syndrome &
  \multicolumn{1}{c|}{1} &
  craniosynostosis - lambdoid - unilateral &
  \multicolumn{1}{c|}{Craniosynostosis} &
  1 \\ \midrule
townes syndrome &
  Syndrome &
  \multicolumn{1}{c|}{1} &
  cleft lip + palate - unilateral - incomplete &
  \multicolumn{1}{c|}{Cleft} &
  1 \\ \midrule
poland syndrome &
  Syndrome &
  \multicolumn{1}{c|}{1} &
  cleft palate, not classified &
  \multicolumn{1}{c|}{Cleft} &
  2 \\ \midrule
van der woude syndrome &
  Syndrome &
  \multicolumn{1}{c|}{1} &
  cleft palate - complete &
  \multicolumn{1}{c|}{Cleft} &
  4 \\ \midrule
crouzon syndrome &
  Syndrome &
  \multicolumn{1}{c|}{4} &
  cleft lip + palate - unilateral - complete &
  \multicolumn{1}{c|}{Cleft} &
  3 \\ \midrule
apert syndrome &
  Syndrome &
  \multicolumn{1}{c|}{7} &
  cleft lip + palate - bilateral - complete &
  \multicolumn{1}{c|}{Cleft} &
  1 \\ \midrule
aarskog syndrome &
  Syndrome &
  \multicolumn{1}{c|}{1} &
  cleft lip + palate, not classified &
  \multicolumn{1}{c|}{Cleft} &
  1 \\ \midrule
syndrome, not classified &
  Syndrome &
  \multicolumn{1}{c|}{7} &
  facial cleft, not classified &
  \multicolumn{1}{c|}{Cleft} &
  1 \\ \midrule
hemifacial myohypertrophy condition &
  Facial Asymmetry &
  \multicolumn{1}{c|}{11} &
  Unknown &
  \multicolumn{1}{c|}{Unknown/other} &
  41 \\ \midrule
craniofacial microsomia &
  Facial Asymmetry &
  \multicolumn{1}{c|}{1} &
  achondroplasia &
  \multicolumn{1}{c|}{Unknown/other} &
  2 \\ \midrule
microtia - unilateral &
  Unknown/other &
  \multicolumn{1}{c|}{8} &
  hypohidrotic ectodermal dysplasia &
  \multicolumn{1}{c|}{Unknown/other} &
  1 \\ \midrule
microtia - bilateral &
  Unknown/other &
  \multicolumn{1}{c|}{1} &
  chondrodysplasia punctata &
  \multicolumn{1}{c|}{Unknown/other} &
  1 \\ \midrule
ectrodactyly\_ectodermal\_dysplasia\_clefting\_syndrome &
  Unknown/other &
  \multicolumn{1}{c|}{1} &
   &
  \multicolumn{1}{c|}{} &
   \\ \bottomrule
\end{tabular}%
}
\caption{Breakdown of clinical images collected to evaluate periorbital distance prediction and segmentation models.}
\label{tab:sup_big_bucket}
\end{table}

\begin{table}[]
\centering
\resizebox{\columnwidth}{!}{%
\begin{tabular}{@{}lccccccc@{}}
\toprule
\textbf{Structure}       & MRD 1 & MRD 2 & ICD   & OCD   & ISS   & SSS   & Horiz Fissure \\ \midrule
\multicolumn{1}{r}{STD:} & 0.540 & 0.499 & 2.049 & 1.088 & 0.109 & 0.041 & 1.230         \\ \bottomrule
\end{tabular}%
}
\caption{Variance for eye periorbital distances calculated between 5 human graders. 100 healthy images were used to determine variance of measurements.}
\label{tab:eye_std}
\end{table}

\begin{table}[]
\centering
\resizebox{\columnwidth}{!}{%
\begin{tabular}{@{}lcccccc@{}}
\toprule
\textbf{Structure}       & Sup. Medial & Sup. Central & Sup. Lateral & Inf. Medial & Inf. Central & Inf. Lateral \\ \midrule
\multicolumn{1}{r}{STD:} & 4.664       & 2.995        & 9.476        & 4.481       & 2.636        & 9.415        \\ \bottomrule
\end{tabular}%
}
\caption{Variance for brow periorbital distances calculated between 5 human graders. 100 healthy images were used to determine variance of measurements.}
\label{tab:brow_std}
\end{table}

% Please add the following required packages to your document preamble:
% \usepackage{booktabs}
% \usepackage{multirow}
% \usepackage{graphicx}
% \usepackage[table,xcdraw]{xcolor}
% Beamer presentation requires \usepackage{colortbl} instead of \usepackage[table,xcdraw]{xcolor}
\begin{table}[]
\centering
\resizebox{\columnwidth}{!}{%
\begin{tabular}{@{}|c|c|c|c|c|c|c|c|c|c|c|c|c|c|c|c|@{}}
\toprule
\cellcolor[HTML]{FFFFFF}\textbf{Dataset} &
  Model &
  MRD 1 &
  MRD 2 &
  ICD &
  IPD &
  OCD &
  ISS &
  SSS &
  Vertical Fissure &
  Horizontal Fissure &
  Canthal Tilt &
  Lat Canthal Height &
  Med Canthal Height &
  Vert Dystopia &
  Average \\ \midrule
                               & DLV3 & 4.52 & 5.16 & 3.25 & 5.81 & 6.45 & 5.19 & 0    & 4.52 & 5.19 & 4.52 & 4.52 & 5.16 & 5.16 & 4.57 \\ \cmidrule(l){2-16} 
\multirow{-2}{*}{Craniofacial} & SAM  & 2.6  & 7.1  & 3.9  & 8.39 & 4.52 & 4.55 & 0.65 & 7.74 & 5.81 & 7.74 & 5.81 & 7.1  & 7.1  & 5.62 \\ \midrule
                               & DLV3 & 5.93 & 4.96 & 4.84 & 6.08 & 4.36 & 4    & 0    & 3.75 & 5.32 & 5.32 & 4.59 & 5.44 & 9.07 & 4.90 \\ \cmidrule(l){2-16} 
\multirow{-2}{*}{Healthy}      & SAM  & 6.42 & 4.85 & 6.9  & 1.46 & 7.26 & 4.52 & 1.96 & 5.58 & 4.36 & 0    & 4.84 & 6.17 & 8.84 & 4.86 \\ \midrule
                               & DLV3 & 6.19 & 4.42 & 5.31 & 1.79 & 3.57 & 2.7  & 2.68 & 6.19 & 5.31 & 6.19 & 6.19 & 6.19 & 0.89 & 4.43 \\ \cmidrule(l){2-16} 
\multirow{-2}{*}{TED}          & SAM  & 7.96 & 4.42 & 7.96 & 2.68 & 5.31 & 7.96 & 6.25 & 5.31 & 7.96 & 4.42 & 4.42 & 6.19 & 2.65 & 5.65 \\ \bottomrule
\end{tabular}%
}
\caption{Percentage of Eye measurements outside the 95\% limits of agreement
(mean difference ± 1.96 standard deviations) of bland altman plots for all models on all datasets. Bilateral distances were averaged. Abbreviations can be interpreted as follows: MRD-Margin Reflex Distance, ICD-inner canthal distance, IPD- interpupillary distance, OCD- outer canthal distance, ISS- inferior scleral show, SSS- superior scleral show, Vert Fissure- vertical palpebral fissure, Horiz Fissure- horizontal palpebral fissure, Med Canthal- medial canthal height, Lat Canthal- lateral canthal height. Craniofacial classes combined for brevity.}
\label{tab:supp_eye_ba}
\end{table}

\begin{table}[]
\centering
\resizebox{\columnwidth}{!}{%
\begin{tabular}{@{}|c|c|c|c|c|c|c|c|c|@{}}
\toprule
\cellcolor[HTML]{FFFFFF}\textbf{Dataset} & Model & Inf. Central & Inf. Lateral & Inf. Medial & Sup. Central & Sup. Lateral & Sup. Medial & Average \\ \midrule
                               & DLV3 & 7.1  & 8.39 & 9.03  & 6.45 & 8.39 & 9.68  & 6.85 \\ \cmidrule(l){2-9} 
\multirow{-2}{*}{Craniofacial} & SAM  & 3.25 & 5.81 & 9.03  & 3.25 & 3.9  & 9.68  & 6.05 \\ \midrule
                               & DLV3 & 3.27 & 0.24 & 5.02  & 4.6  & 0.36 & 4.9   & 4.24 \\ \cmidrule(l){2-9} 
\multirow{-2}{*}{Healthy}      & SAM  & 4.13 & 0.36 & 8.67  & 6.67 & 0.12 & 8.67  & 5.33 \\ \midrule
                               & DLV3 & 5.36 & 2.7  & 10.62 & 4.46 & 2.7  & 10.62 & 5.42 \\ \cmidrule(l){2-9} 
\multirow{-2}{*}{TED}          & SAM  & 6.19 & 4.46 & 7.08  & 4.46 & 3.57 & 7.08  & 5.18 \\ \bottomrule
\end{tabular}%
}
\caption{Percentage of brow measurements outside the 95\% limits of agreement
(mean difference ± 1.96 standard deviations) of bland altman plots for all models on all datasets. Bilateral distances were averaged. Abbreviations can be interpreted as follows: Sup. Medial- superior medial brow height, Sup. Central- superior central brow height, Sup. Lateral- superior lateral brow height, Inf. Medial- inferior medial brow height, Inf. Central- inferior central brow height, Inf. Lateral- inferior lateral brow height. Craniofacial classes combined for brevity.}
\label{tab:supp_brow_ba}
\end{table}

% Please add the following required packages to your document preamble:
% \usepackage{booktabs}
% \usepackage{multirow}
% \usepackage{graphicx}
% \usepackage[table,xcdraw]{xcolor}
% Beamer presentation requires \usepackage{colortbl} instead of \usepackage[table,xcdraw]{xcolor}
\begin{table}[]
\centering
\resizebox{\columnwidth}{!}{%
\begin{tabular}{@{}|c|c|c|c|c|c|c|c|@{}}
\toprule
\cellcolor[HTML]{FFFFFF}\textbf{} &
   &
  Sup. Medial &
  Sup. Central &
  Sup. Lateral &
  Inf. Medial &
  Inf. Central &
  Inf. Lateral \\ \midrule
 &
  DLV3 &
  \textbf{4.23 ± 11.95} &
  \textbf{1.03 ± 0.80} &
  \textbf{1.12 ± 2.11} &
  \textbf{3.95 ± 12.74} &
  \textbf{0.67 ± 0.56} &
  \textbf{1.55 ± 2.30} \\ \cmidrule(l){2-8} 
\multirow{-2}{*}{Healthy} &
  SAM &
  5.65 ± 12.91 &
  1.22 ± 1.15 &
  1.48 ± 3.00 &
  5.74 ± 13.80 &
  0.92 ± 0.74 &
  2.09 ± 3.26 \\ \midrule
 &
  DLV3 &
  \textbf{10.41 ± 18.47} &
  \textbf{3.94 ± 10.39} &
  \textbf{4.71 ± 13.23} &
  \textbf{10.44 ± 19.64} &
  \textbf{3.65 ± 10.84} &
  \textbf{4.71 ± 13.55} \\ \cmidrule(l){2-8} 
\multirow{-2}{*}{TED} &
  SAM &
  12.74 ± 18.50 &
  4.45 ± 8.35 &
  5.96 ± 11.40 &
  13.18 ± 20.39 &
  3.98 ± 8.89 &
  5.41 ± 12.25 \\ \midrule
 &
  DLV3 &
  23.23 ± 30.38 &
  8.7 ± 9.81 &
  2.54 ± 3.97 &
  24.87 ± 32.63 &
  8.88 ± 10.98 &
  \textbf{1.72 ± 2.59} \\ \cmidrule(l){2-8} 
\multirow{-2}{*}{Cleft Palate} &
  SAM &
  \textbf{18.39 ± 21.31} &
  \textbf{4.08 ± 8.11} &
  \textbf{2.03 ± 1.56} &
  \textbf{18.59 ± 23.31} &
  \textbf{3.71 ± 8.92} &
  2.25 ± 1.31 \\ \midrule
 &
  DLV3 &
  14.37 ± 22.94 &
  \textbf{3.14 ± 5.62} &
  10.23 ± 21.32 &
  15.15 ± 24.41 &
  \textbf{2.96 ± 6.06} &
  11.0 ± 22.26 \\ \cmidrule(l){2-8} 
\multirow{-2}{*}{Craniosynostosis} &
  SAM &
  \textbf{9.41 ± 16.4} &
  6.16 ± 15.55 &
  \textbf{1.78 ± 1.14} &
  \textbf{9.83 ± 18.36} &
  6.56 ± 17.83 &
  \textbf{1.56 ± 1.61} \\ \midrule
 &
  DLV3 &
  \textbf{15.79 ± 17.21} &
  4.66 ± 13.46 &
  7.6 ± 21.49 &
  \textbf{16.34 ± 18.39} &
  4.72 ± 14.31 &
  7.87 ± 22.77 \\ \cmidrule(l){2-8} 
\multirow{-2}{*}{Facial Asymmetry} &
  SAM &
  16.26 ± 22.11 &
  \textbf{1.0 ± 0.76} &
  \textbf{1.33 ± 1.77} &
  17.4 ± 23.67 &
  \textbf{1.04 ± 1.11} &
  \textbf{2.66 ± 2.02} \\ \midrule
 &
  DLV3 &
  \textbf{13.32 ± 18.87} &
  6.88 ± 13.89 &
  8.31 ± 15.95 &
  \textbf{13.59 ± 19.93} &
  6.67 ± 14.34 &
  8.67 ± 16.6 \\ \cmidrule(l){2-8} 
\multirow{-2}{*}{Syndrome} &
  SAM &
  14.35 ± 18.18 &
  \textbf{2.9 ± 5.42} &
  \textbf{6.69 ± 13.82} &
  15.3 ± 20.11 &
  \textbf{2.13 ± 5.58} &
  \textbf{6.49 ± 14.29} \\ \bottomrule
\end{tabular}%
}
\caption{Mean Absolute Error (MAE) of all models on all datasets used in this
study for brow measurements. MAE was calculated according to Equation 2, and is reported as +/- the standard deviation. Bold indicates the lowest MAE for each measurement for each model. Bilateral distances were averaged. Abbreviations can be interpreted as follows: Sup. Medial- superior medial brow height, Sup. Central- superior central brow height, Sup. Lateral- superior lateral brow height, Inf. Medial- inferior medial brow height, Inf. Central- inferior central brow height, Inf. Lateral- inferior lateral brow height.}
\label{tab:supp_brow_mae_o}
\end{table}

% Please add the following required packages to your document preamble:
% \usepackage{booktabs}
% \usepackage{multirow}
% \usepackage{graphicx}
% \usepackage[table,xcdraw]{xcolor}
% Beamer presentation requires \usepackage{colortbl} instead of \usepackage[table,xcdraw]{xcolor}
\begin{table}[]
\centering
\resizebox{\columnwidth}{!}{%
\begin{tabular}{@{}|c|c|c|c|c|c|c|c|@{}}
\toprule
\cellcolor[HTML]{FFFFFF}Disease   Bucket &
  Segmentation Model &
  Sup. Medial &
  Sup. Central &
  Sup. Lateral &
  Inf. Medial &
  Inf. Central &
  Inf. Lateral \\ \midrule
 &
  SAM &
  1.68 ± 1.49 &
  \textbf{0.83 ± 0.58} &
  \textbf{1.23 ± 0.96} &
  1.47 ± 1.13 &
  0.74 ± 0.39 &
  1.80 ± 1.22 \\ \cmidrule(l){2-8} 
\multirow{-2}{*}{Healthy} &
  DLV3 &
  \textbf{1.40 ± 1.27} &
  0.90 ± 0.68 &
  1.61 ± 1.00 &
  \textbf{0.75 ± 0.73} &
  \textbf{0.59 ± 0.42} &
  \textbf{1.24 ± 0.93} \\ \midrule
 &
  SAM &
  5.81 ± 7.74 &
  2.54 ± 2.32 &
  3.23 ± 2.63 &
  5.54 ± 8.60 &
  1.86 ± 1.83 &
  2.40 ± 1.83 \\ \cmidrule(l){2-8} 
\multirow{-2}{*}{TED} &
  DLV3 &
  \textbf{2.18 ± 4.60} &
  \textbf{0.90 ± 0.78} &
  \textbf{1.20 ± 1.42} &
  \textbf{1.42 ± 3.49} &
  \textbf{0.89 ± 0.91} &
  \textbf{1.20 ± 1.25} \\ \midrule
 &
  SAM &
  \textbf{1.09 ± 0.48} &
  1.6 ± 0.18 &
  \textbf{0.4 ± 0.11} &
  \textbf{1.11 ± 0.05} &
  1.58 ± 0.62 &
  2.02 ± 0.42 \\ \cmidrule(l){2-8} 
\multirow{-2}{*}{Cleft Palate} &
  DLV3 &
  2.24 ± 3.90 &
  \textbf{0.78 ± 0.76} &
  0.47 ± 0.46 &
  2.29 ± 3.29 &
  \textbf{0.66 ± 0.49} &
  \textbf{1.15 ± 0.56} \\ \midrule
 &
  SAM &
  \textbf{2.83 ± 1.5} &
  0.89 ± 0.34 &
  1.08 ± 0.75 &
  \textbf{2.95 ± 0.46} &
  0.87 ± 0.4 &
  3.05 ± 4.1 \\ \cmidrule(l){2-8} 
\multirow{-2}{*}{Craniosynostosis} &
  DLV3 &
  6.01 ± 10.40 &
  \textbf{0.55 ± 0.57} &
  \textbf{0.89 ± 0.42} &
  6.97 ± 12.35 &
  \textbf{0.56 ± 0.25} &
  \textbf{1.25 ± 1.44} \\ \midrule
 &
  SAM &
  2.65 ± 1.39 &
  \textbf{0.69 ± 0.02} &
  \textbf{0.83 ± 0.09} &
  2.85 ± 1.11 &
  \textbf{0.51 ± 0.04} &
  1.11 ± 0.13 \\ \cmidrule(l){2-8} 
\multirow{-2}{*}{Facial Asymmetry} &
  DLV3 &
  \textbf{2.20 ± 3.04} &
  1.00 ± 0.77 &
  0.90 ± 0.59 &
  \textbf{2.09 ± 3.86} &
  0.97 ± 0.57 &
  \textbf{0.84 ± 0.73} \\ \midrule
 &
  SAM &
  \textbf{1.86 ± 1.8} &
  1.38 ± 1.1 &
  1.59 ± 1.34 &
  \textbf{1.77 ± 1.22} &
  \textbf{0.88 ± 0.56} &
  1.61 ± 0.7 \\ \cmidrule(l){2-8} 
\multirow{-2}{*}{Syndrome} &
  DLV3 &
  2.25 ± 5.46 &
  \textbf{0.91 ± 0.85} &
  \textbf{1.08 ± 1.49} &
  1.78 ± 4.50 &
  1.16 ± 1.57 &
  \textbf{1.36 ± 1.38} \\ \midrule
 &
  SAM &
  1.96 ± 1.58 &
  1.31 ± 0.96 &
  1.4 ± 0.77 &
  1.79 ± 1.27 &
  \textbf{0.93 ± 0.34} &
  1.55 ± 0.7 \\ \cmidrule(l){2-8} 
\multirow{-2}{*}{Unknown/Other Craniofacial} &
  DLV3 &
  \textbf{0.97 ± 1.34} &
  \textbf{1.00 ± 0.74} &
  \textbf{1.25 ± 0.96} &
  \textbf{0.99 ± 1.18} &
  0.98 ± 0.72 &
  \textbf{1.34 ± 1.26} \\ \bottomrule
\end{tabular}%
}
\caption{Mean Absolute Error (MAE) of all models on all datasets used in this study for brow measurements with outliers excluded. Outliers denote measurements that were greater than 1 standard deviation above the MAE (Equation 2). All measurements are reported as +/- the standard deviation. Bilateral distances were averaged. Bold indicates the lowest MAE for each measurement for each model. Abbreviations can be interpreted as follows: Sup. Medial- superior medial brow height, Sup. Central- superior central brow height, Sup. Lateral- superior lateral brow height, Inf. Medial- inferior medial brow height, Inf. Central-inferior central brow height, Inf. Lateral- inferior lateral brow height.}
\label{tab:supp_brow_mae}
\end{table}

% Please add the following required packages to your document preamble:
% \usepackage{booktabs}
% \usepackage{multirow}
% \usepackage{graphicx}
% \usepackage[table,xcdraw]{xcolor}
% Beamer presentation requires \usepackage{colortbl} instead of \usepackage[table,xcdraw]{xcolor}
\begin{table}[]
\centering
\resizebox{\columnwidth}{!}{%
\begin{tabular}{@{}|c|c|cc|cc|@{}}
\toprule
\cellcolor[HTML]{FFFFFF}\textbf{} &
  \textbf{} &
  \multicolumn{2}{c|}{\textbf{Right}} &
  \multicolumn{2}{c|}{\textbf{Left}} \\ \midrule
\textbf{} &
  \textbf{} &
  \multicolumn{1}{c|}{\textbf{Sup. Lateral}} &
  \textbf{Sup. Medial} &
  \multicolumn{1}{c|}{\textbf{Sup. Lateral}} &
  \textbf{Sup. Medial} \\ \midrule
 &
  \textbf{\% Dataset} &
  \multicolumn{2}{c|}{0.85} &
  \multicolumn{2}{c|}{0.86} \\ \cmidrule(l){2-6} 
 &
  \textbf{periorbitAI} &
  \multicolumn{1}{c|}{2.87 ± 2.32} &
  6.68 ± 14.02 &
  \multicolumn{1}{c|}{2.87 ± 2.49} &
  7.04 ± 17.92 \\ \cmidrule(l){2-6} 
 &
  \textbf{SAM} &
  \multicolumn{1}{c|}{1.34 ± 2.09} &
  5.15 ± 14.79 &
  \multicolumn{1}{c|}{1.65 ± 5.78} &
  5.91 ± 18.85 \\ \cmidrule(l){2-6} 
\multirow{-4}{*}{\textbf{Healthy}} &
  \textbf{DLV3} &
  \multicolumn{1}{c|}{\textbf{1.04 ± 0.93}} &
  \textbf{4.12 ± 12.99} &
  \multicolumn{1}{c|}{\textbf{1.06 ± 0.97}} &
  \textbf{4.41 ± 16.16} \\ \midrule
 &
  \textbf{\% Dataset} &
  \multicolumn{2}{c|}{0.58} &
  \multicolumn{2}{c|}{0.63} \\ \cmidrule(l){2-6} 
 &
  \textbf{periorbitAI} &
  \multicolumn{1}{c|}{4.19 ± 8.18} &
  6.46 ± 15.18 &
  \multicolumn{1}{c|}{5.16 ± 12.02} &
  9.63 ± 20.53 \\ \cmidrule(l){2-6} 
 &
  \textbf{SAM} &
  \multicolumn{1}{c|}{4.27 ± 6.05} &
  10.24 ± 18.31 &
  \multicolumn{1}{c|}{\textbf{3.55 ± 10.00}} &
  \textbf{8.74 ± 20.53} \\ \cmidrule(l){2-6} 
\multirow{-4}{*}{\textbf{TED}} &
  \textbf{DLV3} &
  \multicolumn{1}{c|}{\textbf{2.18 ± 5.63}} &
  \textbf{4.59 ± 11.75} &
  \multicolumn{1}{c|}{4.34 ± 16.55} &
  9.51 ± 23.51 \\ \midrule
 &
  \textbf{\% Dataset} &
  \multicolumn{2}{c|}{0.69} &
  \multicolumn{2}{c|}{0.66} \\ \cmidrule(l){2-6} 
 &
  \textbf{periorbitAI} &
  \multicolumn{1}{c|}{12.79  ± 15.46} &
  13.64  ± 17.23 &
  \multicolumn{1}{c|}{14.49    ± 21.16} &
  25.88  ± 35.60 \\ \cmidrule(l){2-6} 
 &
  \textbf{SAM} &
  \multicolumn{1}{c|}{3.88  ± 7.55} &
  \textbf{8.52  ± 15.68} &
  \multicolumn{1}{c|}{10.71    ± 28.57} &
  20.96  ± 32.50 \\ \cmidrule(l){2-6} 
\multirow{-4}{*}{\textbf{Syndrome}} &
  \textbf{DLV3} &
  \multicolumn{1}{c|}{\textbf{2.93  ± 6.99}} &
  15.33  ± 24.24 &
  \multicolumn{1}{c|}{\textbf{9.04    ± 24.78}} &
  \textbf{11.38  ± 24.75} \\ \midrule
 &
  \textbf{\% Dataset} &
  \multicolumn{2}{c|}{1} &
  \multicolumn{2}{c|}{0.82} \\ \cmidrule(l){2-6} 
 &
  \textbf{periorbitAI} &
  \multicolumn{1}{c|}{1.96  ± 2.34} &
  \textbf{2.39  ± 2.02} &
  \multicolumn{1}{c|}{7.41    ± 10.27} &
  20.39  ± 31.52 \\ \cmidrule(l){2-6} 
 &
  \textbf{SAM} &
  \multicolumn{1}{c|}{\textbf{1.17  ± 1.63}} &
  11.59  ± 21.71 &
  \multicolumn{1}{c|}{\textbf{1.57    ± 2.13}} &
  23.23  ± 35.05 \\ \cmidrule(l){2-6} 
\multirow{-4}{*}{\textbf{Facial Asymmetry}} &
  \textbf{DLV3} &
  \multicolumn{1}{c|}{5.82  ± 13.96} &
  17.91  ± 27.97 &
  \multicolumn{1}{c|}{10.53    ± 30.94} &
  \textbf{15.96  ± 31.60} \\ \midrule
 &
  \textbf{\% Dataset} &
  \multicolumn{2}{c|}{0.88} &
  \multicolumn{2}{c|}{1} \\ \cmidrule(l){2-6} 
 &
  \textbf{periorbitAI} &
  \multicolumn{1}{c|}{5.57  ± 5.20} &
  16.05  ± 25.31 &
  \multicolumn{1}{c|}{6.95    ± 5.57} &
  5.21  ± 5.98 \\ \cmidrule(l){2-6} 
 &
  \textbf{SAM} &
  \multicolumn{1}{c|}{1.34  ± 1.01} &
  9.49  ± 19.79 &
  \multicolumn{1}{c|}{\textbf{1.23    ± 0.78}} &
  \textbf{2.66  ± 1.68} \\ \cmidrule(l){2-6} 
\multirow{-4}{*}{\textbf{Craniosynostosis}} &
  \textbf{DLV3} &
  \multicolumn{1}{c|}{\textbf{1.26  ± 1.23}} &
  \textbf{1.59  ± 1.32} &
  \multicolumn{1}{c|}{11.60    ± 30.73} &
  10.39  ± 26.83 \\ \midrule
 &
  \textbf{\% Dataset} &
  \multicolumn{2}{c|}{0.8} &
  \multicolumn{2}{c|}{0.7} \\ \cmidrule(l){2-6} 
 &
  \textbf{periorbitAI} &
  \multicolumn{1}{c|}{5.61  ± 10.04} &
  19.82  ± 23.68 &
  \multicolumn{1}{c|}{11.00    ± 11.89} &
  29.30  ± 41.34 \\ \cmidrule(l){2-6} 
 &
  \textbf{SAM} &
  \multicolumn{1}{c|}{\textbf{2.13  ± 2.05}} &
  \textbf{16.48  ± 21.27} &
  \multicolumn{1}{c|}{\textbf{1.96    ± 1.58}} &
  \textbf{21.58  ± 30.76} \\ \cmidrule(l){2-6} 
\multirow{-4}{*}{\textbf{Cleft}} &
  \textbf{DLV3} &
  \multicolumn{1}{c|}{2.88  ± 4.80} &
  23.83  ± 28.51 &
  \multicolumn{1}{c|}{2.48    ± 4.16} &
  24.09  ± 37.63 \\ \midrule
 &
  \textbf{\% Dataset} &
  \multicolumn{2}{c|}{0.84} &
  \multicolumn{2}{c|}{0.77} \\ \cmidrule(l){2-6} 
 &
  \textbf{periorbitAI} &
  \multicolumn{1}{c|}{8.79 ± 8.98} &
  14.32 ± 20.95 &
  \multicolumn{1}{c|}{12.19 ± 21.11} &
  \textbf{10.41 ± 17.59} \\ \cmidrule(l){2-6} 
 &
  \textbf{SAM} &
  \multicolumn{1}{c|}{\textbf{2.51 ± 6.43}} &
  10.72 ± 22.07 &
  \multicolumn{1}{c|}{6.31 ± 21.56} &
  17.23 ± 32.31 \\ \cmidrule(l){2-6} 
\multirow{-4}{*}{\textbf{Unknown/Other Craniofacial}} &
  \textbf{DLV3} &
  \multicolumn{1}{c|}{3.34 ± 10.21} &
  \textbf{5.26 ± 15.06} &
  \multicolumn{1}{c|}{\textbf{6.29 ± 23.11}} &
  11.21 ± 27.88 \\ \bottomrule
\end{tabular}%
}
\caption{Comparison of MAE (computed using Equation \ref{mae}) of our models to
PeriorbitAI for brow measurements. For all measurements, for both our models and PeriorbitAI, MAE was computed using only images successfully analyzed by PeriorbitAI. ‘\% Dataset’ denotes the percentage of the original dataset for each measurement successfully processed by PeriorbitAI. Bold denotes lowest MAE of each measurement for each dataset. Abbreviations can be interpreted as follows: Sup. Medial- superior medial brow height, Sup. Central- superior central brow height, Sup. Lateral- superior lateral brow height, Inf. Medial- inferior medial brow height, Inf. Central- inferior central brow height, Inf. Lateral- inferior lateral brow height.}
\label{tab:supp_brow_pai}
\end{table}

\begin{table}[]
\centering
\resizebox{\columnwidth}{!}{%
\begin{tabular}{@{}|c|c|c|c|c|c|c|@{}}
\toprule
\cellcolor[HTML]{FFFFFF}\textbf{Dataset} & CAP & Goldenhar & Healthy Aduly & Ptosis & TED & Treacher Collins \\ \midrule
ID         & 0.107 & 0.05 & 0.04 & 0.06 & 0.04 & 0.07 \\ \midrule
OOD        & 0.104 & 0.18 & 0.52 & 0.24 & 0.16 & 0.15 \\ \midrule
Difference & 0.003 & 0.13 & 0.48 & 0.18 & 0.12 & 0.08 \\ \bottomrule
\end{tabular}%
}
\caption{Wasserstein distances between ID train set and ID test set or OOD dataset.}
\label{tab:wasserstein}
\end{table}

\begin{table}[]
\centering
\resizebox{\columnwidth}{!}{%
\begin{tabular}{@{}|c|c|c|c|c|c|c|c|@{}}
\toprule
\cellcolor[HTML]{FFFFFF}\textbf{Dataset} & Classification & Pediatric & Accuracy & Recall & PPV & F1 Score & AUROC \\ \midrule
                      &                         & +  & 0.77 & 0.75 & 0.77 & 0.76 & 0.97 {[}.96-.98{]} \\ \cmidrule(l){3-8} 
                      & \multirow{-2}{*}{XGBoost}   & -  & 0.66 & 0.62 & 0.67 & 0.64 & 0.94 {[}.93-.95{]} \\ \cmidrule(l){2-8} 
                      &                         & +  & 0.66 & 0.63 & 0.67 & 0.65 & 0.96 {[}.95-.96{]} \\ \cmidrule(l){3-8} 
                      & \multirow{-2}{*}{Lasso} & -  & 0.54 & 0.54 & 0.52 & 0.51 & 0.89 {[}.87-.91{]} \\ \cmidrule(l){2-8} 
\multirow{-5}{*}{ID}  & CNN                     & NA & 0.78 & 0.78 & 0.80 & 0.78 & 0.96 {[}.95-.97{]} \\ \midrule
                      &                         & +  & 0.63 & 0.58 & 0.87 & 0.69 & 0.91 {[}.86-.95{]} \\ \cmidrule(l){3-8} 
                      & \multirow{-2}{*}{XGBoost}   & -  & 0.54 & 0.29 & 0.66 & 0.56 & 0.87 {[}.82-.92{]} \\ \cmidrule(l){2-8} 
                      &                         & +  & 0.68 & 0.64 & 0.71 & 0.67 & 0.93 {[}.89-.96{]} \\ \cmidrule(l){3-8} 
                      & \multirow{-2}{*}{Lasso} & -  & 0.57 & 0.57 & 0.61 & 0.57 & 0.87 {[}.82-.92{]} \\ \cmidrule(l){2-8} 
\multirow{-5}{*}{OOD} & CNN                     & NA & 0.13 & 0.13 & 0.29 & 0.18 & 0.62 {[}.56-.68{]} \\ \bottomrule
\end{tabular}%
}
\caption{Influence of removing pediatric variable from machine learning classifiers when performing disease classification. }
\label{tab:ped-ablation}
\end{table}

\begin{table}[]
\centering
\resizebox{\columnwidth}{!}{%
\begin{tabular}{@{}|c|c|c|c|@{}}
\toprule
\cellcolor[HTML]{FFFFFF}\textbf{Feature} & \textbf{\% Importance Lasso} & \textbf{Feature} & \textbf{\% Importance XGBoost} \\ \midrule
IPD                  & 0.10 & ISS Difference      & 0.06 \\ \midrule
ICD                  & 0.06 & IPD                 & 0.05 \\ \midrule
L. MRD 2             & 0.05 & SSS Difference      & 0.04 \\ \midrule
Right Horiz. Fissure & 0.04 & L. Horiz. Fissure   & 0.04 \\ \midrule
L. Med. Canthal Ht.  & 0.04 & ICD                 & 0.04 \\ \midrule
L. Scleral Area      & 0.04 & OCD                 & 0.04 \\ \midrule
Vert. Dystopia       & 0.04 & R. Horiz Fissure    & 0.04 \\ \midrule
L. MRD 1             & 0.04 & R. MRD 1            & 0.04 \\ \midrule
R. MRD 1             & 0.04 & R. Med. Canthal. Ht & 0.04 \\ \midrule
L. Lat. Canthal Ht.  & 0.04 & Vert. Dystopia      & 0.03 \\ \bottomrule
\end{tabular}%
}
\caption{Feature importances of models trained without influence of pediatric status.}
\label{tab:feat-import}
\end{table}

\end{document}